%% file: iclr2026_conference.tex
\documentclass{article} % For LaTeX2e
\usepackage[T1]{fontenc}
\usepackage{iclr2026_conference,times}

\input{math_commands.tex}

\usepackage{algorithm}
\usepackage{algorithmic}
\usepackage{booktabs}
\usepackage{caption}
\usepackage{graphicx}
\usepackage{multirow}
\usepackage{nicematrix}
\usepackage{pgfplots}
\pgfplotsset{compat=1.16}
\usepackage{hyperref}
\usepackage{subcaption}
\usepackage{url}
\usepackage{wrapfig}
\usepackage[x11names]{xcolor}
\usepackage{tcolorbox}
\newtheorem{theorem}{Theorem}[section]

\tcbuselibrary{breakable}

\title{Thinking on the Fly: Test-Time Reasoning\\Enhancement via Latent Thought\\Policy Optimization}

\author{%
Wengao Ye$^{1}$\thanks{Correspondence to: Wengao Ye <\texttt{wengao.ye@kellogg.ox.ac.uk}>},~~~Yan Liang$^{2}$,~~~Lianlei Shan$^{3}$\\
$^{1}$University of Oxford, $^{2}$University College London, $^{3}$University of Chinese Academy of Sciences
}

\iclrfinalcopy % Uncomment for camera-ready version, but NOT for submission.
\begin{document}

\maketitle

\begin{abstract}
Recent advancements in Large Language Models (LLMs) have shifted from explicit Chain-of-Thought (CoT) reasoning to more efficient latent reasoning, where intermediate thoughts are represented as vectors rather than text. However, latent reasoning can be brittle on challenging, out-of-distribution tasks where robust reasoning is most critical. To overcome these limitations, we introduce Latent Thought Policy Optimization (LTPO), a parameter-free framework that enhances LLM reasoning entirely at test time, without requiring model parameter updates. LTPO treats intermediate latent ``thought'' vectors as dynamic parameters that are actively optimized for each problem instance. It employs an online policy gradient method guided by an intrinsic, confidence-based reward signal computed directly from the frozen LLM's own output distributions, eliminating the need for external supervision or expensive text generation during optimization. Extensive experiments on five reasoning benchmarks show that LTPO not only matches or surpasses strong baselines on standard tasks but also demonstrates remarkable robustness where others fail. Most notably, on highly challenging AIME benchmarks where existing latent reasoning baselines collapse to near-zero accuracy, LTPO delivers substantial improvements, showcasing a unique capability for complex reasoning.\footnote{Source code: \url{https://github.com/ltpo2025/LTPO}}
\end{abstract}

\section{Introduction}

Large Language Models (LLMs) have dramatically advanced the state of reasoning in artificial intelligence, propelled by \textbf{Chain-of-Thought (CoT)} prompting \citep{wei2022cot}, which enhances performance by decomposing complex problems into explicit intermediate reasoning steps in natural language. Despite its effectiveness, the explicit generation of CoT is often inefficient and costly, both in computational resources and inference latency. To overcome this, recent research has focused on \emph{latent reasoning}, which encodes the intermediate ``thoughts'' as continuous hidden vectors within the model's latent space. Notable approaches such as Coconut \citep{hao2024training} and SoftCoT \citep{xu2025softcot} have demonstrated that by replacing textual CoT with learned latent embeddings, models can attain similar reasoning accuracy with increased computational efficiency.

However, a major limitation of existing latent reasoning approaches is their brittleness when handling challenging, out-of-distribution tasks. Methods relying on offline trained projection or latent modules typically fail to generalize to hard problems or new domains in practice, as shown in our experiments, advanced latent reasoning baselines can completely collapse on competition-grade benchmarks where robust reasoning is most critical.

In this work, we propose \textbf{Latent Thought Policy Optimization (LTPO)}, a parameter-free and entirely test-time framework for boosting LLM reasoning capabilities. Unlike prior latent reasoning approaches that fix the intermediate representations, LTPO treats the \emph{latent thought vectors} as dynamic parameters to be actively optimized at test time. Concretely, given a prompt augmented with $K$ special \emph{latent thought tokens}, we iteratively refine their hidden vectors in a closed-loop process: at each step, the current latent states are perturbed, passed through the frozen LLM (no model weight updates), and evaluated by an \emph{intrinsic confidence-based reward} which is computed directly from the model's output distributions, without textual decoding. This reward signal guides a policy gradient update, improving the latent thoughts to maximize model's predictive certainty. After a small number of optimization steps, the refined latent thought vectors are concatenated with the prompt's embeddings, and passed through the LLM to generate the final answer.

LTPO is simple, general, and requires no additional data or model parameter updates, enabling \emph{on-the-fly} enhancement of LLM reasoning abilities, purely via test-time latent space optimization. Extensive experiments on five mathematical reasoning benchmarks demonstrate that LTPO not only matches or surpasses strong baselines on standard tasks, but remains robust where others fail. Most notably, on highly challenging AIME competition benchmarks where existing latent reasoning methods collapse to near-zero accuracy, LTPO delivers substantial improvements: for instance, with Qwen-2.5-7B-Instruct \citep{yang2024qwen2technicalreport}, LTPO achieves 16.67\% and 13.33\% accuracy on AIME2024 and AIME2025 respectively, dramatically outperforming all competitive baselines.

Our key contributions are summarized as follows:
\begin{itemize}
\setlength\itemsep{0em}
    \item We propose LTPO, a test-time framework for LLM reasoning, which directly optimizes intermediate latent thought vectors via policy gradient, without any model fine-tuning.
    \item We introduce an efficient online Reinforcement Learning (RL) loop using a self-derived confidence-based reward, computed from the LLM’s own output distributions, eliminating the need for extra supervision or expensive text generation during optimization.
    \item We empirically show that LTPO delivers robust reasoning performance on diverse and challenging benchmarks, addressing critical failure points of previous latent reasoning approaches.
\end{itemize}

%% Please note that we have introduced automatic line number generation
%% into the style file for \LaTeXe. This is to help reviewers
%% refer to specific lines of the paper when they make their comments. Please do
%% NOT refer to these line numbers in your paper as they will be removed from the
%% style file for the final version of accepted papers.

\section{Related Work}
\label{related_work}

The pursuit of advanced reasoning in LLMs has evolved from generating explicit textual rationales, known as CoT prompting \citep{wei2022cot,Kojima2022zeroshot,zhou2023leasttomost}, to more efficient latent reasoning approaches. Latent reasoning represents intermediate ``thoughts'' as continuous vectors in the model's hidden space rather than as discrete text, aiming to match the performance of explicit CoT with lower latency. Most methods in this area rely on an offline training phase to learn these latent representations. For example, iCoT \citep{deng2024icot} fine-tunes a LLM on problems and their detailed reasoning steps to generate both the reasoning and final answer in a single forward pass; Coconut \citep{hao2024training} directly trains the model to use the last hidden state of the ``thought token'' as the next input embedding; Similarly, SoftCoT \citep{xu2025softcot} utilizes a lightweight assistant model to generate soft thought tokens that are then projected into the LLM’s latent space using a trained projection module. Although improving efficiency, these approaches can be brittle, especially on challenging, out-of-distribution problems where adaptability is most needed.

A parallel line of research has applied RL to enhance LLMs. The dominant paradigm, originating with Reinforcement Learning from Human Feedback (RLHF) \citep{ouyang2022rlhf,bai2022constitutionalaiharmlessnessai}, involves fine-tuning the model's parameters using algorithms like PPO \citep{schulman2017ppo} to align with human preferences. This approach has been extended by more efficient methods like DPO \citep{Rafailov2023dpo} and specialized techniques for reasoning like GRPO \citep{shao2024grpo,deepseekai2025deepseekr1} and DAPO \citep{yu2025dapo}. However, these methods universally operate by updating the model's weights in an offline training stage, based on external reward signals.

Our method, LTPO, diverges from both of these established research directions. It operates at the intersection of latent reasoning and optimization but introduces a new paradigm: enhancing reasoning entirely at test time without updating any model parameters. This approach is situated within the growing interest in test-time optimization strategies, which have recently been explored for instance-level adaption at test time \citep{li2025latentseek}. However, LTPO operationalizes this concept through a distinct mechanism: it establishes a self-contained RL loop where the reward signal is derived directly from the LLM’s own output distributions. This bypasses the need for explicit text generation or external evaluation within the RL loop, allowing for a more effective and efficient optimization of the latent reasoning path.

\begin{figure}[t]
%\vskip -0.3in
\centering
\includegraphics[width=1.0\textwidth]{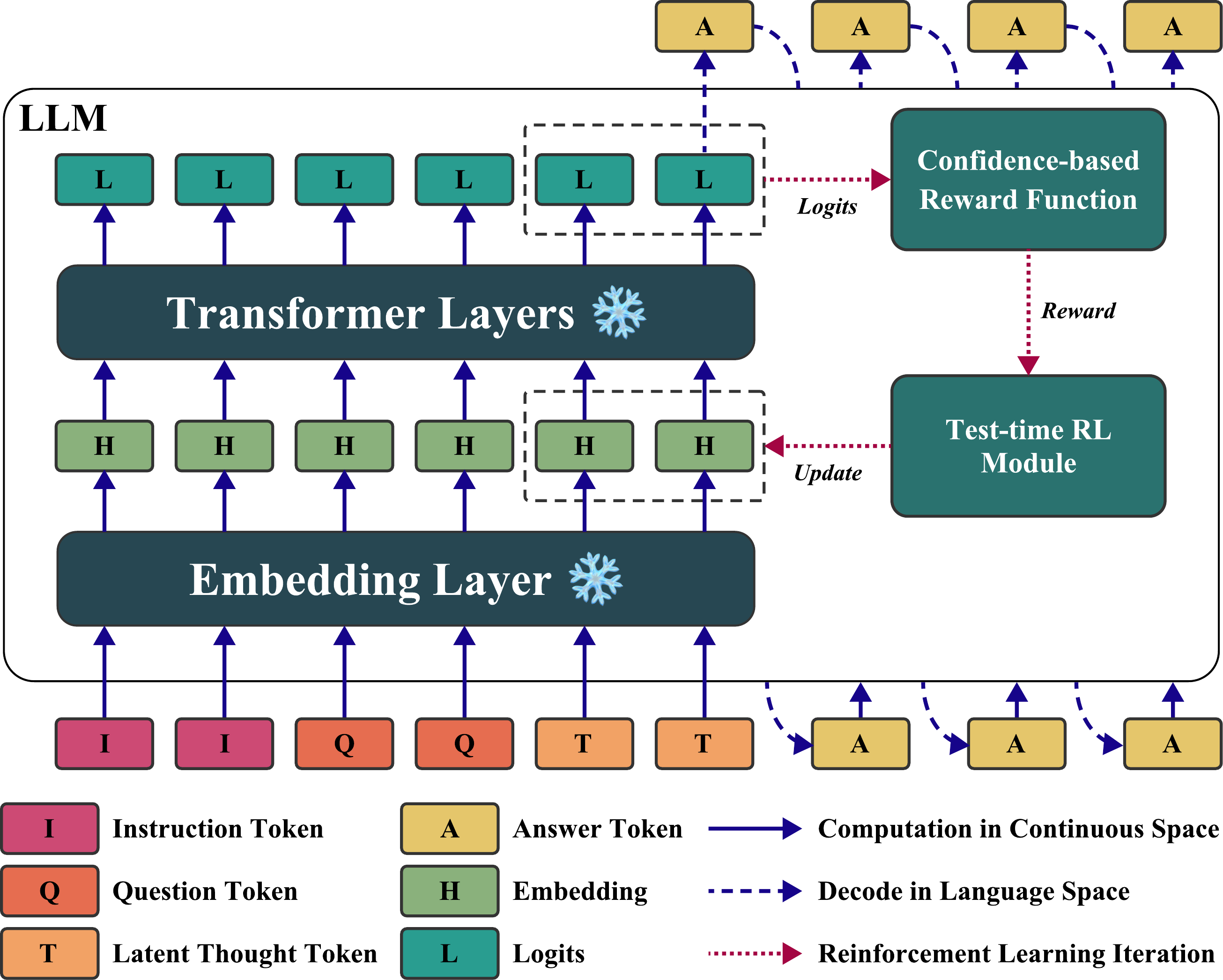}
\caption{Overview of LTPO. The framework iteratively refines the embedding vectors of the \emph{latent thought tokens} via a test-time RL loop. A confidence-based reward, calculated from the LLM’s output logits, guides the Test-time RL Module to update the latent thought vectors. After optimization, the refined vectors are concatenated with the prompt's embeddings and passed through the LLM to generate the final answer.}
\label{fig:overview}
\end{figure}

\section{Method}
\label{method}

We introduce \textbf{LTPO}, a parameter-free framework that enhances the reasoning abilities of LLMs at test time. Figure~\ref{fig:overview} shows the overview of our framework.

\subsection{Latent Thought Representation}

Let $\gM_{\vtheta}$ denote a LLM with frozen parameters $\vtheta$. Given a standard input prompt which is a sequence of discrete tokens $\vx = (x_1, ..., x_N)$, the model's embedding layer $E$ maps it to a sequence of vectors $E(\vx) = (\ve_1, ..., \ve_N)$, where $\ve_i \in \sR^d$ and $d$ is the model's hidden dimension.

To enable latent reasoning, we augment the input prompt with $K$ special placeholder tokens, which we call \emph{latent thought tokens}. Their initial embedding vectors, denoted by $\mH^{(0)} = (\vh_1^{(0)}, ..., \vh_K^{(0)}) \in \sR^{K \times d}$, are obtained by passing the placeholder tokens (e.g., \texttt{[THINK]}) through the model's embedding layer:
\begin{align}
    (\vh_1^{(0)}, ..., \vh_K^{(0)}) = E(\mathrm{[THINK]}_1, ..., \mathrm{[THINK]}_K).\label{eq:init}
\end{align}

These latent thought vectors $\mH$ serve as the parameters that will be optimized at test time. The full input to the LLM at any point is the concatenation of the original prompt embeddings and the current latent thought vectors, $E(\vx) \mathbin\Vert \mH$.

\subsection{Optimization via Test-Time Reinforcement Learning}

We formulate the problem of finding the optimal latent thought vectors $\mH$ as a sequential decision-making task, which we solve using a direct policy gradient method over $T$ timesteps.

\paragraph{Policy as latent space exploration.} The optimization process iteratively refines the latent thought vectors $\mH$, which themselves serve as the parameters of our policy. We define the components of this process as follows:

\textit{State:} The state at timestep $t$ is the current set of latent thought vectors, $\mH^{(t)} = (\vh_1^{(t)}, ..., \vh_K^{(t)})$.

\textit{Action:} An action $\mA^{(t)}$ is a candidate for the next state of the latent thoughts. The action space is continuous, where $\mA^{(t)} = (\va_1^{(t)}, ..., \va_K^{(t)})$ and $\va_k^{(t)} \in \sR^d$.

\textit{Policy:} We employ a simple yet effective stochastic policy that explores the latent space by sampling actions from a multivariate Gaussian distribution centered at the current state $\mH^{(t)}$:
\begin{align}
    \mA^{(t)} \sim \pi(\cdot | \mH^{(t)}) = \gN(\mH^{(t)}, \vsigma^2 \mI),\label{eq:policy}
\end{align}

where $\vsigma^2$ is a variance hyperparameter that controls exploration and is decayed over time. Sampling an action is equivalent to applying a random perturbation, $\mA^{(t)} = \mH^{(t)} + \veps^{(t)}$, where $\veps^{(t)} \sim \gN(\vzero, \vsigma^2 \mI)$. This encourages local exploration around promising areas of the latent thought space.

\paragraph{Confidence-based reward.} To guide the optimization without external supervision, we introduce an intrinsic reward signal derived directly from the LLM's own output distributions. Unlike recent works that leverage confidence for post-hoc selection \citep{kang2025selfcertainty,fu2025deepthinkconf}, our application is distinct. LTPO employs confidence as a dense reward signal to actively guide an online optimization process within the latent space for each problem instance. The intuition is that a well-formed reasoning path should lead the model to be more confident in its subsequent predictions. While this intrinsic signal is powerful, we acknowledge that model confidence is not a perfect proxy for correctness, a potential limitation we discuss in Appendix~\ref{appendix:confidence_vs_correctness}.

Given a candidate action $\mA^{(t)}$ (a probe of next state), we compute the reward by first passing it through the frozen LLM $\gM_{\vtheta}$ along with the prompt embeddings $E(\vx)$ to obtain output probability distributions over the vocabulary $\mV$ for each individual token:
\begin{align}
    (P_1^{(t)}, ..., P_{|\vx| + K}^{(t)}) = \mathrm{softmax}(\gM_{\vtheta}(E(\vx) \mathbin\Vert \mA^{(t)})).\label{eq:probability}
\end{align}

We define the confidence for a single latent thought vector $\va^{(t)}$ as the mean negative log-probability of the top-$k$ most probable tokens predicted from that position:
\begin{align}
    C(\va_i^{(t)}) = - \frac{1}{k} \sum_{v \in \mathrm{top}(k, P_i^{(t)})} \log P_i^{(t)}(v),\label{eq:confidence}
\end{align}

where $\mathrm{top}(k, P_i^{(t)})$ is the set of the $k$ most probable tokens in the distribution $P_i^{(t)}$. A higher confidence score indicates the model is more certain about its prediction. The final reward for the entire sequence of latent thoughts is the average confidence across all $K$ vectors:
\begin{align}
    R(\mA^{(t)}) = \frac{1}{K} \sum_{i=1}^{K} C(\va_i^{(t)}).\label{eq:reward}
\end{align}

\paragraph{Latent thought update with policy gradient.} Since the reward function $R(\cdot)$ is non-differentiable with respect to the input latent vectors, we cannot use standard backpropagation. Instead, we can use a direct policy gradient method, such as REINFORCE~\citep{williams1992simple}, to update the policy parameters (i.e., the latent vectors). The objective is to find latent thought vectors $\mH$ that maximize the expected reward:
\begin{align}
    J(\mH) = \E_{\mA \sim \pi(\cdot | \mH)}[R(\mA)].\label{eq:objective}
\end{align}

The gradient of this objective is given by the policy gradient theorem:
\begin{align}
    \nabla_{\mH} J(\mH) = \E_{\mA \sim \pi(\cdot | \mH)}[R(\mA) \nabla_{\mH} \log \pi(\mA | \mH)].\label{eq:policy_gradient}
\end{align}

For our Gaussian policy (Equation~\ref{eq:policy}), the log-probability gradient term can be simplified to:
\begin{align}
    \nabla_{\mH} \log \pi(\mA | \mH) = \frac{(\mH + \veps) - \mH}{\vsigma^2} = \frac{\veps}{\vsigma^2}.\label{eq:gradient}
\end{align}

This yields a Monte Carlo estimate of the gradient from a single sample $\veps^{(t)}$ at each optimization step $t$:
\begin{align}
    \nabla_{\mH} J(\mH^{(t)}) \approx R(\mH^{(t)} + \veps^{(t)})\frac{\veps^{(t)}}{\vsigma^2}.\label{eq:monte_carlo}
\end{align}

The latent thought vectors are then updated via standard gradient ascent:
\begin{align}
    \mH^{(t+1)} = \mH^{(t)} + \eta \cdot R(\mH^{(t)} + \veps^{(t)})\frac{\veps^{(t)}}{\vsigma^2},\label{eq:grad_ascent}
\end{align}

where $\eta$ is the learning rate. This update rule intuitively pushes the latent thoughts $\mH$ in the direction that yields higher rewards.

After $T$ optimization steps, the set of the optimized latent thought vectors $\mH^\ast$ is concatenated with the prompt embeddings and passed through the LLM to autoregressively generate the final answer:
\begin{align}
    \vy = \mathrm{Decoder}(\gM_{\vtheta}(E(\vx) \mathbin\Vert \mH^\ast)).\label{eq:decode}
\end{align}

\section{Experiments}
\label{experiments}

In this section, we present our experimental setup, results, and analysis to validate the effectiveness of LTPO. Addtional implementation details are provided in Appendix~\ref{appendix:implementation}.

\subsection{Experimental Setup}
\label{experimental_setup}

\paragraph{Models.} We evaluate our method on four open-source reasoning LLMs from two distinct model families: LLaMA-3.1-8B-Instruct, LLaMA-3.2-3B-Instruct \citep{grattafiori2024llama3herdmodels}, Qwen-2.5-7B-Instruct \citep{yang2024qwen2technicalreport}, and Qwen-3-14B \citep{yang2025qwen3technicalreport}. These models are fully open-source for reproducibility, covering multiple parameter scales to test robustness and generalizability.

\paragraph{Datasets.} Following \cite{deng2024icot} and \cite{liu20251bllmsurpass405b}, we focus on mathematical reasoning for evaluation. We conduct experiments on five datasets: GSM8K \citep{cobbe2021gsm8k}, MATH-500 \citep{hendrycks2021math500}, ASDiv \citep{miao2020asdiv}, AIME2024 \citep{aime2024}, and AIME2025 \citep{aime2025}. GSM8K and MATH-500 are commonly used for evaluating the performance of reasoning systems, thus providing a solid basis for comparative analysis. ASDiv is a benchmark consisting of varied, multi-step elementary school math word problems. AIME2024 and AIME2025 are high-difficulty mathematical competition problems which are widely adopted in recent evaluations of top reasoning LLMs (e.g., Gemini-2.5 \citep{google2025gemini}, Grok-4 \citep{xai2025grok4}, GPT-5 \citep{openai2025gpt5}) and featured in the MathArena leaderboard \citep{balunović2025matharena}. Specifically, we utilize the augmented version of ASDiv (ASDiv-Aug) from \cite{xu2025softcot} to ensure that the models encounter novel instances, given that LLaMA and Qwen are well-trained LLMs.

\paragraph{Baselines.} We compare LTPO against three strong and directly comparable baselines to provide a thorough evaluation of its effectiveness: (1) \textbf{Zero-Shot CoT}: Our foundational baseline is the zero-shot CoT prompting method from \cite{Kojima2022zeroshot}. This approach represents the standard for eliciting reasoning in the \emph{discrete token space} by instructing the model to generate explicit, step-by-step thinking. On modern LLMs, this technique sets a very strong performance baseline, often exceeding fine-tuned models, making it an essential point of comparison. (2) \textbf{Zero-Shot CoT-Unk}: We directly append untuned \texttt{[UNK]} tokens to the prompt to perform CoT reasoning. This baseline isolates the contribution of our test-time optimization procedure for latent thought tokens. (3) \textbf{SoftCoT}: To provide a more direct and advanced comparison for our latent reasoning framework, we adopt SoftCoT \citep{xu2025softcot}, which performs reasoning in the \emph{continuous latent space}. It has been shown to outperform methods like LoRA fine-tuning \citep{hu2022lora} and Coconut \citep{hao2024training} on various datasets. (4) \textbf{LatentSeek}: We also compare against LatentSeek \citep{li2025latentseek}, which applies instance-level policy gradient optimization in the latent space at test time. Unlike our approach, LatentSeek relies on full autoregressive decoding to evaluate each intermediate step. Including this baseline allows us to highlight the critical efficiency and robustness advantages of our intrinsic confidence-based reward over computationally intensive decoding-based optimization methods.

\paragraph{Evaluation.} All results are reported in accuracy (\%). For each problem, we perform greedy decoding to generate the final answer. The generated output is then parsed to extract the final numerical answer, which is compared against the ground truth.

% In Section~\ref{broader_comparison}, we conduct a further analysis comparing LTPO against other state-of-the-art paradigms, including RL-based methods, to highlight its unique advantages.

\begin{table}[t]
\caption{Performance of LTPO vs. baselines across four models and five reasoning benchmarks, reported in accuracy (\%). The optimal results are in bold and the suboptimal ones are underlined.}
\label{tab:main_results}
\centering
\small
\begin{NiceTabular}{llccccc|c}
\toprule
\multirow{2}{*}{\textbf{Model}} & \multirow{2}{*}{\textbf{Method}} & \multirow{2}{*}{\textbf{GSM8K}} & \textbf{MATH} & \textbf{ASDiv} & \textbf{AIME} & \textbf{AIME} & \multirow{2}{*}{\textbf{Avg.}} \\ & & & \textbf{500} & \textbf{Aug} & \textbf{2024} & \textbf{2025} & \\ \midrule
\midrule

\multirow{6}{*}{\begin{tabular}[c]{@{}l@{}}LLaMA-3.1-8B-\\Instruct\end{tabular}}
& Zero-Shot CoT & 76.88 & 48.60 & 79.58 & 6.67 & 0.00 & 42.35 \\
& Zero-Shot CoT-Unk & 80.67 & 46.20 & 89.50 & \underline{10.00} & \underline{3.33} & \underline{45.94} \\
& SoftCoT & 80.36 & 39.80 & 87.57 & 0.00 & 0.00 & 41.55 \\
& LatentSeek (Prompt 1) & 53.15 & \textbf{52.20} & 46.15 & 3.33 & 0.00 & 30.97 \\
& LatentSeek (Prompt 2) & \textbf{83.70} & 47.40 & \textbf{90.66} & 3.33 & 0.00 & 45.02 \\
& \textbf{LTPO (Ours)} & \underline{81.27} & \underline{49.00} & \underline{89.69} & \textbf{16.67} & \textbf{6.67} & \textbf{48.66} \\ \midrule

\multirow{4}{*}{\begin{tabular}[c]{@{}l@{}}LLaMA-3.2-3B-\\Instruct\end{tabular}}
& Zero-Shot CoT & 69.60 & \underline{45.60} & 85.84 & \underline{6.67} & 0.00 & 41.54 \\
& Zero-Shot CoT-Unk & \underline{76.65} & 45.00 & \underline{87.82} & 0.00 & \underline{3.33} & \underline{42.56} \\
& SoftCoT & 71.49 & 35.80 & 85.74 & 0.00 & 0.00 & 38.61 \\
& \textbf{LTPO (Ours)} & \textbf{76.88} & \textbf{46.40} & \textbf{88.63} & \textbf{13.33} & \textbf{6.67} & \textbf{46.38} \\ \midrule

\multirow{6}{*}{\begin{tabular}[c]{@{}l@{}}Qwen-2.5-7B-\\Instruct\end{tabular}}
& Zero-Shot CoT & \underline{88.55} & \underline{72.40} & 91.71 & 10.00 & \underline{10.00} & \underline{54.53} \\
& Zero-Shot CoT-Unk & 86.20 & 70.80 & 91.52 & \underline{13.33} & \underline{10.00} & 54.37 \\
& SoftCoT & 85.29 & 65.20 & 88.92 & 0.00 & 0.00 & 47.88 \\
& LatentSeek (Prompt 1) & \textbf{89.54} & \textbf{73.60} & 91.43 & 10.00 & 0.00 & 52.91 \\
& LatentSeek (Prompt 2) & 86.20 & 57.80 & \underline{91.81} & 6.67 & 6.67 & 49.83 \\
& \textbf{LTPO (Ours)} & 88.17 & \textbf{73.60} & \textbf{92.20} & \textbf{16.67} & \textbf{13.33} & \textbf{56.79} \\ \midrule

\multirow{4}{*}{\begin{tabular}[c]{@{}l@{}}Qwen-3-14B\end{tabular}}
& Zero-Shot CoT & \underline{91.96} & 68.20 & 92.77 & \textbf{10.00} & \textbf{10.00} & \underline{54.59} \\
& Zero-Shot CoT-Unk & 91.66 & 71.20 & \underline{94.03} & \underline{6.67} & 0.00 & 52.71 \\
& SoftCoT & \textbf{93.78} & \underline{72.60} & 91.62 & 0.00 & 0.00 & 51.60 \\
& \textbf{LTPO (Ours)} & 91.89 & \textbf{73.00} & \textbf{94.41} & \textbf{10.00} & \textbf{10.00} & \textbf{55.86} \\ \bottomrule

\end{NiceTabular}
\end{table}

\subsection{Main Results}

The comprehensive experimental results are presented in Table~\ref{tab:main_results}. Our proposed method, LTPO, demonstrates superior performance across all tested models and benchmarks. A key observation is that LTPO achieves the highest average accuracy in every model category, establishing its effectiveness and generalizability. For instance, with Qwen-2.5-7B-Instruct, LTPO achieves an average accuracy of 56.79\%, surpassing Zero-Shot CoT by 2.26 points, SoftCoT by 8.91 points, and LatentSeek by 3.88 points. Similarly, with LLaMA-3.1-8B-Instruct, LTPO scores 48.66\% on average, outperforming Zero-Shot CoT by 6.31 points, SoftCoT by 7.11 points, and LatentSeek by 3.64 points. These results strongly support that LTPO is an effective strategy for enhancing LLM reasoning. A deeper analysis of the results reveals several key strengths of the LTPO framework:

\paragraph{Exceptional robustness on challenging problems.} The most notable result is LTPO's performance on the highly difficult AIME2024 and AIME2025 benchmarks. As hypothesized, these problems highlight the brittleness of existing latent reasoning methods. The advanced SoftCoT baseline completely collapses on these tasks, scoring 0.00\% on AIME2024 and AIME2025 for all tested models. This result likely stems from the fact that SoftCoT's static projection module, which is trained on GSM8K data, does not generalize to the new domain of AIME competition mathematics. This highlights a key failure mode for trained latent reasoning methods when faced with out-of-distribution problems. While LatentSeek demonstrates competitive performance on simpler tasks like GSM8K, it also struggles to generalize to these competition-grade problems. In stark contrast, LTPO demonstrates remarkable resilience. With Qwen-2.5-7B-Instruct, LTPO achieves 16.67\% on AIME2024 and 13.33\% on AIME2025, dramatically outperforming not only the failed SoftCoT baseline but also the strong Zero-Shot CoT (10.00\% on both). This unique capability to make substantial gains on problems where other methods fail underscores the value of our on-the-fly optimization process for navigating complex reasoning spaces. To further investigate the scalability of LTPO under a high-compute regime, we evaluate Qwen-3-14B on AIME benchmarks with the maximum generation length extended from 4,096 to 64,000 tokens in Section~\ref{extended_gen_len}.

\paragraph{Effectiveness of test-time optimization.} By comparing LTPO to the Zero-Shot CoT-Unk baseline, we can isolate the contribution of our optimization procedure. The Zero-Shot CoT-Unk method, which uses untuned latent thought tokens, underperforms LTPO for nearly all cases. For example, on LLaMA-3.2-3B-Instruct, LTPO's average accuracy of 46.38\% is 3.82 points higher than Zero-Shot CoT-Unk's 42.56\%. Similarly, on Qwen-3-14B, LTPO's average accuracy is 3.15 points higher than Zero-Shot CoT-Unk's 52.71\%. This margin confirms that the performance gains are not merely from adding placeholder tokens but are fundamentally driven by the optimization of the latent thought vectors at test time.

\paragraph{Generalizability across models.} The consistent superiority of LTPO across both LLaMA and Qwen model families, and at scales from 3B to 14B parameters, highlights the generalizability of our approach. LTPO does not rely on any model specific architecture or pretraining objective. Instead, it leverages the model’s intrinsic confidence signal, a universal property of probabilistic models, making it broadly applicable across LLMs.

\begin{table}[t]
\caption{Performance of LTPO vs. training-based baselines on GSM8K, MATH-500, and AIME-2024 with Llama-3.1-8B-Instruct, reported in accuracy (\%). The ``Train Model Params'' column indicates whether the method requires updating model parameters. The symbol $\dagger$ indicates the accuracy is reported by \cite{zeng2025simplerlzoo}. The optimal results are in bold and the suboptimal ones are underlined.}
\label{tab:broader_results}
\centering
\begin{NiceTabular}{l|cc|ccc}
\toprule
\multirow{2}{*}{\textbf{Method}} & \textbf{Train Model} & \textbf{Supervision} & \multirow{2}{*}{\textbf{GSM8K}} & \textbf{MATH} & \textbf{AIME} \\
& \textbf{Params} & \textbf{Type} & & \textbf{500} & \textbf{2024} \\ \midrule
\midrule

Genius & Yes & Self & 78.09 & \underline{47.60} & \underline{3.33} \\
SimpleRL-Zoo$^{\dagger}$ & Yes & External Data & 79.20 & 23.00 & 0.00 \\
iCoT & Yes & External Data & 50.05 & - & - \\
SoftCoT & Yes & External Data & \underline{80.36} & 39.80 & 0.00 \\
\textbf{LTPO (Ours)} & No & Self & \textbf{81.27} & \textbf{49.00} & \textbf{16.67} \\ \bottomrule

\end{NiceTabular}
\end{table}

\subsection{Comparison with Training-based Methods}
\label{broader_comparison}

\begin{figure}[t]
%\vskip -0.3in
\centering

\begin{subfigure}{0.5\textwidth}
\centering
\begin{tikzpicture}[scale=0.8]
\definecolor{GREEN}{RGB}{82,157,63}
\definecolor{RED}{RGB}{198,58,51}
\definecolor{BLUE}{RGB}{57,119,175}
\definecolor{ORANGE}{RGB}{240,133,54}
\begin{axis}[
    xlabel={Number of Thought Tokens},
    ylabel={Accuracy (\%)},
    xmin=1, xmax=16,
    ymin=60, ymax=90,
    xtick={1, 2, 4, 8, 12, 16},
    ytick={60, 65, 70, 75, 80, 85, 90},
    legend pos=south west,
    ymajorgrids=true,
]

\addplot[
    color=RED,
    style=dashed,
    line width=0.75mm,
    ]
    coordinates {
    (1,79.58)(2,79.58)(4,79.58)(8,79.58)(12,79.58)(16,79.58)
    };
    \addlegendentry{Zero-Shot CoT}
\addplot[
    color=BLUE,
    mark=triangle*,
    line width=0.75mm,
    ]
    coordinates {
    (1,83.04)(2,87.19)(4,85.16)(8,71.58)(12,74.76)(16,62.52)
    };
    \addlegendentry{SoftCoT}
\addplot[
    color=GREEN,
    mark=square*,
    line width=0.75mm,
    ]
    coordinates {
    (1,88.73)(2,89.69)(4,87.76)(8,88.82)(12,88.44)(16,88.34)
    };
    \addlegendentry{LTPO (Ours)}

\end{axis}
\end{tikzpicture}
\end{subfigure}%
\begin{subfigure}{0.5\textwidth}
\centering
\begin{tikzpicture}[scale=0.8]
\definecolor{GREEN}{RGB}{82,157,63}
\definecolor{RED}{RGB}{198,58,51}
\definecolor{BLUE}{RGB}{57,119,175}
\definecolor{ORANGE}{RGB}{240,133,54}
\definecolor{PURPLE}{RGB}{175,45,185}
\definecolor{AZURE}{RGB}{85,188,190}
\begin{axis}[
    xlabel={Number of Iterations},
    ylabel={Accuracy (\%)},
    xmin=5, xmax=40,
    ymin=87.0, ymax=89.2,
    xtick={5, 10, 20, 30, 40},
    ytick={87.0, 87.5, 88.0, 88.5, 89.0, 89.5, 90.0},
    legend pos=south west,
    ymajorgrids=true,
]

\addplot[
    color=AZURE,
    mark=*,
    line width=0.75mm,
    ]
    coordinates {
    (5,88.63)(10,88.82)(20,89.11)(30,88.54)(40,88.25)
    };
    \addlegendentry{w/ best reward}
\addplot[
    color=PURPLE,
    mark=diamond*,
    line width=0.75mm,
    ]
    coordinates {
    (5,88.25)(10,88.73)(20,88.63)(30,87.66)(40,87.09)
    };
    \addlegendentry{w/o best reward}

\end{axis}
\end{tikzpicture}
\end{subfigure}

\caption{Left: The impact of thought token numbers. Right: The impact of LTPO using thought tokens with best reward. Both are tested on ASDiv-Aug using LLaMA-3.1-8B-Instruct.}
\label{fig:linecharts}
\end{figure}
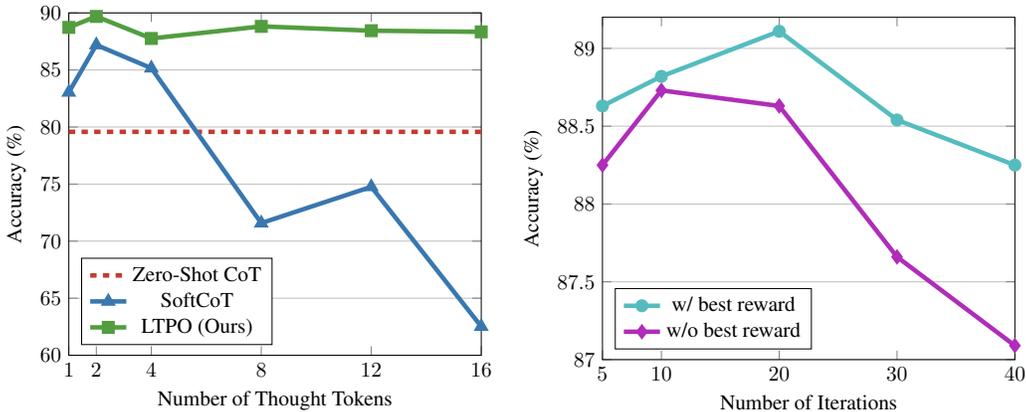

To further situate LTPO within the landscape of reasoning methods, we conduct a targeted comparison against different paradigms on the LLaMA-3.1-8B-Instruct model. We introduce three additional baselines that require a model training phase and/or external training data: (1) \textbf{Genius} \citep{xu2025genius} fine-tunes the model using a \emph{self-reward} mechanism derived from the model's own reasoning process. (2) \textbf{SimpleRL-Zoo} \citep{zeng2025simplerlzoo} fine-tunes the model using a \emph{verifiable-reward} mechanism derived from external data. (3) \textbf{iCoT} \citep{deng2024icot} fine-tunes the model to internalize CoT steps for generating the reasoing and final answer in a single forward pass.

As shown in Table~\ref{tab:broader_results}, LTPO outperforms all baselines on Llama-3.1-8B-Instruct. Notably, LTPO achieves these results without any updates to the model's parameters, operating entirely at test time. This stands in stark contrast to RL-based methods like Genius and SimpleRL-Zoo, which require a costly offline training phase. On GSM8K, LTPO's score of 81.27\% surpasses both Genius (78.09\%) and SimpleRL-Zoo (79.20\%). While LTPO (49.00\%) also leads on MATH-500, outperforming Genius (47.60\%) and more than doubling the score of SimpleRL-Zoo (23.00\%), its advantage is even more stark on the highly challenging AIME2024 benchmark. Here, LTPO achieves 16.67\% accuracy, dramatically outperforming both Genius (3.33\%) and SimpleRL-Zoo (0.00\%). These results demonstrate that our test-time optimization of latent thoughts offers a more effective and robust alternative to offline policy alignment via RL.

\subsection{Analysis of Inference Efficiency}
\label{inference_efficiency}

While test-time optimization can introduce inference overhead, LTPO is designed for efficiency. To quantify this, we benchmark its average inference time against baselines on GSM8K, AIME2024, and AIME2025 using the LLaMA-3.1-8B-Instruct model. We set the number of optimization steps to $T=20$ for LTPO, a value that yields strong accuracy (as shown in Figure~\ref{fig:linecharts}, right).

\begin{wraptable}{r}{0.6\textwidth}
\caption{Average inference time per problem on GSM8K, AIME2024, and AIME2025 with LLaMA-3.1-8B-Instruct, reported in seconds. We set the RL steps $T = 20$ for LTPO, and $T = 10$ for LatentSeek to follow the setting in the original paper. Results are averaged on 3 runs per benchmark per method. The optimal results are in bold and the suboptimal ones are underlined. The ``P1'' and ``P2'' represent ``Prompt 1'' and ``Prompt 2'', respectively.}
\label{tab:run_time}
\centering
\small
\begin{NiceTabular}{l|ccc|c}
\toprule
\multirow{2}{*}{\textbf{Method}} & \multirow{2}{*}{\textbf{GSM8K}} & \textbf{AIME} & \textbf{AIME} & \multirow{2}{*}{\textbf{Avg.}} \\
& & \textbf{2024} & \textbf{2025} & \\ \midrule
\midrule

Zero-Shot CoT & \underline{5.81} & 62.59 & 55.60 & 41.33 \\
SoftCoT & 5.91 & \textbf{26.01} & \underline{35.03} & \underline{22.32} \\
LatentSeek (P1) & 48.99 & 872.47 & 708.40 & 543.29 \\
LatentSeek (P2) & 61.83 & 1002.30 & 512.43 & 525.52 \\
LTPO ($T=20$)  & \textbf{5.69} & \underline{31.80} & \textbf{24.89} & \textbf{20.79} \\ \bottomrule

\end{NiceTabular}
\end{wraptable}

The results in Table~\ref{tab:run_time} show that: (1) on the simpler GSM8K benchmark, LTPO, SoftCoT, and Zero-Shot CoT have nearly identical inference times (about 5.7–5.9 seconds), indicating the overhead of the LTPO loop is not substantial. In stark contrast, LatentSeek requires significantly more time (48.99–61.83 seconds), which is an order of magnitude slower. (2) on the challenging AIME benchmarks, which require much longer reasoning chains, the explicit Zero-Shot CoT method is slow, yet LatentSeek is prohibitively expensive, taking up to 1000 seconds per problem (LatentSeek with prompt 2 on AIME2024). Crucially, LTPO is not only faster than Zero-Shot CoT but is also computationally competitive with SoftCoT, achieving the lowest average inference time overall (20.79 seconds vs. over 500 seconds for LatentSeek).

LTPO’s efficiency stems from a crucial difference in how computation is used: (1) Standard CoT methods rely on autoregressive decoding to generate a lengthy textual chain of thought. (2) LatentSeek relies on full text generation at each intermediate step during the optimization, leading to excessive latency. In contrast, the $T$ forward passes within the LTPO optimization loop do not involve any autoregressive decoding steps; instead, they are computationally efficient passes on a fixed-length input (prompt + latent tokens) to calculate the reward, completely bypassing text generation within the optimization loop. (3) SoftCoT requires an assistant model to first generate soft thought tokens before the main model can process the problem and produce an answer, while LTPO operates within a single model, avoiding the multi-stage overhead of text generation. The final answer in the LTPO framework is decoded only once after the latent thought vectors have been optimized. Consequently, LTPO's entire test-time procedure is less expensive than Zero-Shot CoT and SoftCoT, and over 25 times faster than LatentSeek on average, demonstrating that it enhances reasoning without introducing a substantial latency penalty. We provide a further analysis of optimization overhead and generation savings between Zero-Shot CoT and LTPO in Appendix~\ref{appendix:inference_analysis}.

\subsection{Analysis of Key Components}

We conduct further experiments to analyze the behavior of LTPO and its key components. These studies are performed on the ASDiv-Aug benchmark.

\subsubsection{Impact of the Number of Thought Tokens}
\label{impact_of_num_thought_tokens}

Both LTPO and SoftCoT use special tokens to represent reasoning in latent space. In Figure~\ref{fig:linecharts} (left), we investigate the effect of varying the number of these tokens. We observe that LTPO's performance is stable and robust, remaining high as the number of thought tokens increases from 1 to 16. In contrast, SoftCoT's performance degrades sharply when more than 4 thought tokens are used. This suggests that SoftCoT's static linear projection mechanism struggles to produce high-quality representations for a larger number of thought tokens, whereas LTPO's dynamic test-time optimization can effectively refine them toward an optimal state regardless of their quantity. We provide a further analysis of this stability in Appendix~\ref{appendix:latent_tokens_analysis}.

\subsubsection{Impact of Best Reward Selection}

LTPO iteratively updates the latent thought vectors to maximize a confidence-based reward. A key choice is whether to use the vectors from the final optimization step or the vectors that achieved the highest reward at any point during the process. Figure~\ref{fig:linecharts} (right) shows that using the thought tokens with the best reward consistently outperforms using those from the final iteration. This indicates that the optimization trajectory may not be monotonic, and the final state may not be the optimal one found during exploration. By tracking the highest confidence discovered during exploration, LTPO can more reliably elicit the best possible response from the model, a finding that aligns with similar observations by \cite{kang2025selfcertainty}. Furthermore, the diminishing returns from extended optimization steps point to a potential divergence between the model's predictive confidence and its actual reasoning correctness, a limitation we explore in Appendix~\ref{appendix:confidence_vs_correctness}.

\subsubsection{Sensitivity to the Top-k Reward Hyperparameter}

A critical component of LTPO is the confidence-based reward function (Equation~\ref{eq:confidence}), which relies on a single hyperparameter $k$. We analyze the sensitivity of LTPO to this top-$k$ hyperparameter by varying $k$ from 5 to 100 on ASDiv-Aug, with the results presented in Table~\ref{tab:topk}.

\begin{table}[h]
\caption{The impact of the top-$k$ hyperparameter on LTPO performance, reported in accuracy (\%) on the ASDiv-Aug benchmark.}
\label{tab:topk}
\centering
\begin{NiceTabular}{l|cccccc}
\toprule
Model & $k=5$ & $k=10$ & $k=20$ & $k=50$ & $k=75$ & $k=100$ \\ \midrule
\midrule

LLaMA-3.1-8B-Instruct & 88.92 & \textbf{89.11} & 88.63 & 88.92 & 88.73 & 88.63 \\
LLaMA-3.2-3B-Instruct & \textbf{87.48} & 87.28 & 87.19 & 87.09 & 87.38 & 86.99 \\
Qwen-2.5-7B-Instruct  & \textbf{92.20} & 91.81 & 91.33 & 91.62 & 92.00 & 91.91 \\
\bottomrule

\end{NiceTabular}
\end{table}

\paragraph{High stability and robustness.} The primary observation is the remarkable stability of LTPO's performance across a wide range of $k$ values. For LLaMA-3.1-8B-Instruct, accuracy varies by less than 0.5 percentage points (from 88.63\% to 89.11\%) as $k$ increases from 5 to 100. This pattern holds for all tested models, indicating that the optimization process is not sensitive to the precise value of this hyperparameter.

\paragraph{Optimal $k$ is in the small-to-medium range.} While performance is generally stable, the optimal value for $k$ consistently appears in the smaller range ($k \leq 10$) across all models. This empirically validates that the most potent signal of model confidence is captured by the very highest probability tokens.

\paragraph{Practical implications.} This robustness is a practical advantage of LTPO. It reduces the overhead of hyperparameter tuning and suggests that the core benefit of the method comes from the directional signal of the confidence-based reward rather than its precise formulation. The optimization process appears to successfully capture the gradient of model certainty, whether that certainty is measured over the top 5 or top 100 tokens. This indicates that for a confident model, the probability mass is sufficiently concentrated in the first few tokens, and including more tokens from the long, low-probability tail of the distribution does not introduce significant noise to the reward signal.

\subsection{Generalization to Other Domains}
\label{generalization}

To verify that LTPO generalizes beyond mathematical domains, we evaluate it on StrategyQA \citep{geva2021strategyqa} for commonsense reasoning and Date Understanding \citep{srivastava2023bigbench} from BIG-benchmark for symbolic reasoning.

As detailed in Appendix~\ref{appendix:generalization}, LTPO consistently outperforms the Zero-Shot CoT baseline across both tasks and models. Notably, on the symbolic reasoning task, LTPO with Qwen-2.5-7B-Instruct achieves 76.40\% accuracy, delivering a substantial +12.0\% improvement over the baseline. Crucially, the results highlight the adaptability of test-time optimization compared to fixed latent representations. While the training-based SoftCoT baseline degrades performance on StrategyQA, LTPO successfully improves upon the baseline. This confirms that LTPO is a robust, domain-agnostic framework capable of enhancing reasoning on tasks significantly distinct from standard mathematical benchmarks.

\subsection{Scalability with Extended Generation Length}
\label{extended_gen_len}

\begin{wraptable}{r}{0.5\textwidth}
\caption{Performance of LTPO vs. Zero-Shot CoT on AIME2024 and AIME2025 with Qwen-3-14B, reported in accuracy (\%). The maximum output tokens is set to 64,000. The optimal results are in bold.}
\label{tab:long_reasoning}
\centering
\begin{NiceTabular}{l|cc|c}
\toprule
\multirow{2}{*}{\textbf{Method}} & \textbf{AIME} & \textbf{AIME} & \multirow{2}{*}{\textbf{Avg.}} \\
& \textbf{2024} & \textbf{2025} & \\ \midrule
\midrule

Zero-Shot CoT & 80.00 & 70.00 & 75.00 \\
\textbf{LTPO (Ours)} & \textbf{83.33} & \textbf{76.67} & \textbf{80.00} \\ \bottomrule

\end{NiceTabular}
\end{wraptable}

We further investigate if LTPO remains beneficial when the base model is allowed ample compute for extensive reasoning. We evaluate Qwen-3-14B on AIME benchmarks with the maximum generation length extended to 64k tokens.

As shown in Table~\ref{tab:long_reasoning}, under this high-compute regime, the Zero-Shot CoT baseline improves dramatically to 75.00\% average accuracy (a massive increase from the 10.00\% accuracy observed under standard settings in Table~\ref{tab:main_results}). However, LTPO continues to provide significant gains, boosting performance to 80.00\% (+5.0\%) on average. This result suggests that LTPO is complementary to long-context explicit reasoning: even when the model can ``think'' extensively in natural language, optimizing the initial latent thought vectors provides a better starting point for the reasoning trajectory.

\section{Conclusion}

We introduce LTPO, a parameter-free framework that enhances LLM reasoning entirely at test time by directly optimizing latent thought vectors. LTPO employs an online policy gradient method guided by an intrinsic, confidence-based reward signal to refine its reasoning path for each problem without requiring model parameter updates. This approach overcomes the critical brittleness of existing latent reasoning methods, demonstrating robust reasoning ability on the highly challenging AIME benchmarks where LTPO delivers substantial accuracy gains while advanced latent reasoning baselines collapse to near-zero. Our work thus establishes test-time latent space optimization as a powerful and practical paradigm for eliciting more robust and effective reasoning from LLMs.

\bibliography{iclr2026_conference}
\bibliographystyle{iclr2026_conference}

\clearpage
\appendix

\section{Experimental Details}
\label{appendix:implementation}

This section provides a comprehensive overview of the experimental setup used to evaluate our proposed method, LTPO, ensuring full reproducibility of our results.

\subsection{Computational Resources}

All experiments are conducted on a single A100-80GB or a single RTX4090-48GB GPU. All baseline models, including SoftCoT and iCoT, are trained on a single A100-80GB GPU. For Genius baseline, we directly use the official fine-tuned model~\footnote{\url{https://huggingface.co/xufangzhi/Genius_Magpie-25K_LLaMA3.1-8B-Instruct}} provided by \cite{xu2025genius}.

\subsection{Evaluation Protocol}

Performance is measured by accuracy (\%). For each problem, we use greedy decoding to generate the response and set the maximum output length to 4096 tokens. A parser is used to extract the final numerical answer from the generated text, which is then compared against the ground-truth. A fixed random seed is used across all experiments to ensure consistency.

\subsection{Baseline Implementation Details}

\paragraph{Zero-Shot CoT.} We follow the standard methodology proposed by \cite{Kojima2022zeroshot}. We append the same instruction to every problem prompt: ``Please reason step by step, and put your final answer within \verb|\boxed{}|.''

\paragraph{Zero-Shot CoT-Unk.} This baseline uses the exact same prompt template as LTPO (shown in Appendix~\ref{appendix:ltpo_implemetation}), including the special placeholder tokens for latent thoughts. However, the embeddings for these tokens are not optimized at test time. This setup is designed to isolate the performance gains attributable specifically to our test-time RL optimization loop.

\paragraph{SoftCoT.} We use the official code\footnote{\url{https://github.com/xuyige/SoftCoT}} provided by \cite{xu2025softcot}. For benchmarks that lack a dedicated training set (MATH-500, AIME2024, and AIME2025), we follow the original authors' protocol and use the projection module trained on the GSM8K dataset.

\paragraph{LatentSeek.} We use the official code \footnote{\url{https://github.com/bigai-nlco/LatentSeek}} provided by \cite{li2025latentseek}.

\subsection{LTPO Implementation Details}
\label{appendix:ltpo_implemetation}

\paragraph{Prompt design.} To ensure a fair comparison, LTPO uses the same prompt template as the Zero-Shot CoT-Unk for all experiments. See Appendix~\ref{appendix:prompt_templates} for examples.

\paragraph{Hyperparameter selection protocol.} The performance of LTPO is dependent on several test-time hyperparameters, as detailed in Table~\ref{tab:ltpo_hyperparams}. To determine these values, we employ a grid search methodology. For benchmarks with standard training splits, such as GSM8K and ASDiv-Aug, we hold out a validation set with 200 examples from the training data. For benchmarks without a training set (MATH-500, AIME2024, AIME2025), we use the GSM8K training set as a proxy validation environment to find a reasonable starting point, followed by minimal adjustments. While our framework is training-free, we acknowledge that finding the optimal set of hyperparameters for a new task requires this selection process. This represents a trade-off for the method's adaptability and robustness without requiring any model parameter updates.

To quantify this trade-off and evaluate the robustness of LTPO, we conduct an additional experiment using a single, fixed default hyperparameter configuration for all tasks, removing the need for per-task tuning. The results, shown in Table~\ref{tab:fixed_hyperparams}, reveal that even without task-specific tuning, LTPO remains highly effective.

\begin{table}[t]
\caption{Hyperparameters for LTPO. lr: learning rate.}
\label{tab:ltpo_hyperparams}
\centering
\small
\begin{NiceTabular}{lccccccc}
\toprule
\textbf{model} & \textbf{benchmark} & \textbf{\# thought tokens} & \textbf{\# steps} & \textbf{top-$k$} & \textbf{sigma} & \textbf{sigma decay} & \textbf{lr} \\ \midrule
\midrule

\multirow{5}{*}{\begin{tabular}[c]{@{}@{}l@{}@{}}LLaMA-3.1-8B\\Instruct\end{tabular}}
& GSM8K & 10 & 20 & 10 & 10 & 0.9 & 1e-2 \\
& MATH500 & 10 & 20 & 10 & 5 & 0.9 & 5e-4 \\
& ASDiv-Aug & 2 & 20 & 10 & 10 & 0.9 & 1e-2 \\
& AIME2024 & 8 & 10 & 10 & 4 & 0.9 & 4e-2 \\
& AIME2025 & 8 & 20 & 10 & 5 & 0.9 & 5e-3 \\ \midrule

\multirow{5}{*}{\begin{tabular}[c]{@{}@{}l@{}@{}}LLaMA-3.2-3B\\Instruct\end{tabular}}
& GSM8K & 4 & 20 & 5 & 20 & 0.95 & 5e-3 \\
& MATH500 & 8 & 20 & 20 & 20 & 0.95 & 1e-4 \\
& ASDiv-Aug & 4 & 5 & 10 & 1 & 1.0 & 1e-3 \\
& AIME2024 & 4 & 20 & 10 & 10 & 0.95 & 1e-2 \\
& AIME2025 & 8 & 20 & 10 & 10 & 0.8 & 1e-3 \\ \midrule

\multirow{5}{*}{\begin{tabular}[c]{@{}@{}l@{}@{}}Qwen-2.5-7B\\Instruct\end{tabular}}
& GSM8K & 8 & 20 & 10 & 20 & 0.95 & 1e-2 \\
& MATH500 & 8 & 20 & 10 & 20 & 0.95 & 5e-3 \\
& ASDiv-Aug & 5 & 20 & 5 & 10 & 0.9 & 1e-2 \\
& AIME2024 & 8 & 20 & 10 & 5 & 0.95 & 5e-2 \\
& AIME2025 & 10 & 10 & 10 & 20 & 0.95 & 1e-4 \\ \midrule

\multirow{5}{*}{\begin{tabular}[c]{@{}@{}l@{}@{}}Qwen-3-14B\end{tabular}}
& GSM8K & 10 & 20 & 10 & 20 & 0.95 & 1e-3 \\
& MATH500 & 10 & 20 & 10 & 20 & 0.95 & 5e-3 \\
& ASDiv-Aug & 5 & 20 & 10 & 10 & 0.9 & 1e-2 \\
& AIME2024 & 4 & 20 & 10 & 8 & 0.9 & 4e-4 \\
& AIME2025 & 4 & 20 & 10 & 8 & 0.9 & 1e-2 \\ \bottomrule
\end{NiceTabular}
\end{table}

\begin{table}[h]
\caption{Robustness experiment with a fixed hyperparameter set on LLaMA-3.1-8B-Instruct. Reported in accuracy (\%). The fixed set used is: \# thought tokens = 8, \# steps = 20, top-$k$ = 10, sigma = 5, sigma decay = 0.9, lr = 5e-3. The optimal results are in bold and the suboptimal ones are
underlined.}
\label{tab:fixed_hyperparams}
\centering
\begin{NiceTabular}{lcccccc}
\toprule
\multirow{2}{*}{\textbf{Method}} & \multirow{2}{*}{\textbf{GSM8K}} & \textbf{MATH} & \textbf{ASDiv} & \textbf{AIME} & \textbf{AIME} & \multirow{2}{*}{\textbf{Avg.}} \\ & & \textbf{500} & \textbf{Aug} & \textbf{2024} & \textbf{2025} & \\ \midrule
\midrule

Zero-Shot CoT & 76.88 & 48.60 & 79.58 & 6.67 & 0.00 & 42.35 \\
SoftCoT & 80.36 & 39.80 & 87.57 & 0.00 & 0.00 & 41.55 \\
LatentSeek (Prompt 1) & 49.89 & \textbf{52.20} & 46.15 & 3.33 & 0.00 & 30.31 \\
LatentSeek (Prompt 2) & \textbf{83.70} & 47.40 & \textbf{90.66} & 3.33 & 0.00 & 45.02 \\
LTPO (Per-task Tuned) & \underline{81.27} & \underline{49.00} & \underline{89.69} & \textbf{16.67} & \textbf{6.67} & \textbf{48.66} \\
LTPO (Fixed Default Config) & 80.29 & 45.80 & 88.05 & \underline{13.33} & \textbf{6.67} & \underline{46.83} \\
\bottomrule
\end{NiceTabular}
\end{table}

The LTPO with a fixed hyperparamter set achieves an average accuracy of 46.83\%, only a minor decrease from the 48.66\% achieved with per-task tuning. Most importantly, it still decisively outperforms Zero-Shot CoT (42.35\%), SoftCoT (41.55\%), LatentSeek with prompt 1 (30.31\%), and LatentSeek with prompt 2 (45.02\%), and maintains its unique advantage on the difficult AIME benchmarks. This result strongly suggests that while per-task tuning can provide a final optimization boost, the core reasoning enhancement from LTPO is robust and can be achieved with a general-purpose configuration, affirming its practical utility.

\section{Limitation: The Divergence of Confidence and Correctness}
\label{appendix:confidence_vs_correctness}

\subsection{Qualitative Analysis}
\label{appendix:conf_corr_qualitative}

To better understand the limitations of LTPO, we analyze its failure cases. A key failure mode arises from the nature of the confidence-based reward. In some instances, the optimization process successfully increases the model's confidence, but in a flawed reasoning direction, leading to a confidently incorrect answer.

This confirms that the intrinsic reward, while effective, is not a perfect proxy for correctness and highlights a known challenge for self-improving systems. We illustrate this divergence with a qualitative example from the MATH-500 dataset in Table~\ref{tab:case_study}. The model is tasked with finding the least positive integer multiple of 30 composed only of digits 0 and 2.

The LTPO loop optimizes for the highest reward, which corresponds to the model's internal confidence. In this case, the process converges on an incorrect solution (120) that achieved a higher reward (6.375) than the correct solution (2220), which had a lower reward (4.781).

The incorrect reasoning latches onto a flawed intermediate step (``Step 5: Multiplying 222 by 2 will make it divisible by 10...'') and then attempts to modify it, leading to a series of logical errors. Despite being wrong, the step-by-step structure is fluent and assertive. The LTPO loop refines the latent thoughts to produce this confident, yet ultimately wrong, solution based on the initial error. In contrast, the correct reasoning is sound but produces a lower confidence score. This suggests the optimization can overfit to the model's internal sense of confidence, even when that confidence is misplaced.

Future work could explore incorporating uncertainty estimation or other signals to create more robust rewards that better align with logical correctness.

\subsection{Theoretical Analysis}
\label{appendix:conf_corr_theoretical}

To analyze the relationship between confidence and correctness, it is productive to model the problem geometrically. The space of possible reasoning strategies can be conceptualized as a high-dimensional manifold. On this manifold, we can define two distinct scalar functions: a ``confidence'' function, which is perceptible and optimizable by the LLM at test time, and a ``correctness'' function, which represents the hidden ground truth. This geometric perspective provides a framework for understanding the dynamics of LTPO and the reasons for its successes and failures.

\paragraph{Latent thought manifold.} LTPO optimizes a set of $K$ latent thought vectors, $\mH = (\vh_1, ..., \vh_K)$, where each $\vh_i \in \sR^d$ and $d$ is the model's hidden dimension. The space of all possible sets of these vectors, $\mH \in \sR^{K \times d}$, can be viewed as a high-dimensional manifold, $\gH = \sR^{K \times d}$, where each point represents a unique reasoning strategy.

\paragraph{Confidence landscape.} For any point $\mH$, the model's policy $\pi(\cdot | \mH)$ samples actions (probes) to compute a reward. The objective of LTPO is to maximize the expected reward, which defines the confidence landscape: $J_{conf}(\mH) = \E_{\mA \sim \pi(\cdot | \mH)}[R(\mA)]$. This landscape is observable, and LTPO performs stochastic gradient ascent upon it.

\paragraph{Correctness landscape.} We can define a theoretical function, $f_{corr}(\mA) \rightarrow \{0, 1\}$, which is 1 if an action $\mA$ leads to a correct final answer and 0 otherwise. While $f_{corr}$ itself is discrete, non-differentiable step function, we are interested in the expected correctness: $J_{corr}(\mH) = \E_{\mA \sim \pi(\cdot | \mH)}[f_{corr}(\mA)]$. This expectation is a smooth function of $\mH$, and its gradient, $\nabla_{\mH} J_{corr}(\mH)$, is well-defined via the policy gradient theorem. This landscape represents the ``ground truth'' we ultimately wish to optimize. However, it is fundamentally non-observable at test time, as determining the correctness of an answer requires an external verifier.

\paragraph{Update dynamics.} LTPO navigates the latent thought manifold $\gH$ by iteratively updating the parameters $\mH$ to maximize the confidence objective. At each step $t$, the update follows the gradient of the confidence landscape, defined by the rule $\mH^{(t+1)} = \mH^{(t)} + \eta \nabla_{\mH} J_{conf}(\mH^{(t)})$, where $\eta$ is the learning rate. This dynamic can be modeled as a \emph{gradient flow} in the continuous limit. Specifically, in the limit of a small learning rate $\eta \rightarrow 0$, the trajectory of the parameters $\mH$ follows the \emph{gradient flow} (Ordinary Differential Equation) defined by the expected reward:
\begin{align}
    \frac{d\mH}{dt} = \nabla_{\mH} J_{conf}(\mH).\label{eq:grad_flow}
\end{align}

\begin{theorem}
\label{theorem:optimal_cond}
(The Gradient Alignment Condition). A gradient ascent step on the expected confidence landscape, $J_{conf}(\mH)$, also increases the expected correctness landscape, $J_{corr}(\mH)$, if and only if the gradients of the two landscapes are positively aligned. Formally, this condition is:
\begin{align}
    \nabla_{\mH} J_{conf}(\mH) \cdot \nabla_{\mH} J_{corr}(\mH) > 0.\label{eq:optimal_cond}
\end{align}
\end{theorem}

\emph{Proof:}
\begin{enumerate}
    \item The LTPO update rule moves the latent thought vectors $\mH$ in the direction of the estimated confidence gradient: $\Delta \mH = \eta \cdot \nabla_{\mH} J_{conf}(\mH)$, for a small learning rate $\eta > 0$.

    \item The change in expected correctness from this step can be approximated by the directional derivative of $J_{corr}$ in the direction of the update. For an infinitesimal step (consistent with the \emph{gradient flow} in Equation~\ref{eq:grad_flow}), this change is given by $\Delta J_{corr} \approx \nabla_{\mH} J_{corr}(\mH) \cdot \Delta \mH$.

    \item Substituting the expression for $\Delta \mH$, we get $\Delta J_{corr} \approx \eta(\nabla_{\mH} J_{corr}(\mH) \cdot \nabla_{\mH} J_{conf}(\mH))$.

    \item Since $\eta > 0$, the change in expected correctness, $\Delta J_{corr}$, is positive if and only if the dot product of the two gradients, $\nabla_{\mH} J_{corr}(\mH) \cdot \nabla_{\mH} J_{conf}(\mH)$, is positive. This means the update step is productive (i.e., it increases correctness) only when the angle between the two gradients is acute ($< 90$ degrees).
\end{enumerate}

\begin{theorem}
\label{theorem:conf_incorr_trap}
(The ``Confidently Incorrect'' Trap). Let $\mH_{trap} \in \gH$ be a strict local maximizer of expected confidence landscape $J_{conf}$, such that $J_{corr}(\mH_{trap}) < \epsilon$ for some small $\epsilon > 0$. Let $\gB(\mH_{trap}) \subset \gH$ be the basin of attraction of $\mH_{trap}$ under the gradient flow of $J_{conf}$ (Equation~\ref{eq:grad_flow}). For any initialization $\mH^{(0)} \in \gB(\mH_{trap})$, the trajectory converges to $\mH_{trap}$, resulting in a solution with expected correctness less than $\epsilon$.
\end{theorem}

\emph{Proof:}
\begin{enumerate}
    \item To prove convergence to $\mH_{trap}$, we construct a Lyapunov function. Consider the function $V(\mH) = J_{conf}(\mH_{trap}) - J_{conf}(\mH)$. Inside the neighborhood of $\mH_{trap}$, since $\mH_{trap}$ is a strict local maximizer, $V(\mH) > 0$ for all $\mH \neq \mH_{trap}$ and $V(\mH_{trap}) = 0$.\\\\
    Taking the time derivative of $J_{conf}(\mH)$ along the trajectory of the \emph{gradient flow}:
    \begin{align}
        \frac{d}{dt} J_{conf}(\mH) = \nabla_{\mH} J_{conf}(\mH) \cdot \frac{d\mH}{dt}.
    \end{align}\\
    Substituting the dynamics (Equation~\ref{eq:grad_flow}):
    \begin{align}
        \frac{d}{dt} J_{conf}(\mH) = \lVert \nabla_{\mH} J_{conf}(\mH) \rVert^{2} \geq 0.
    \end{align}\\
    This implies that $J_{conf}(\mH)$ is strictly increasing along any non-stationary trajectory.\\\\
    Consequently, $V(\mH)$ is strictly decreasing within the neighborhood of $\mH_{trap}$. By Lyapunov's Stability Theorem \citep{lyapunov1992stability}, the equilibrium point $\mH_{trap}$ is \emph{asymptotically stable}.

    \item By the definition of the basin of attraction $\gB(\mH_{trap})$, for any initial condition $\mH^{(0)} \in \gB(\mH_{trap})$, the limit of the flow is:
    \begin{align}
        \lim_{t \rightarrow \infty} \mH = \mH_{trap}.
    \end{align}\\
    For the discrete-time algorithm used in LTPO (Equation~\ref{eq:grad_ascent}), standard approximation theory for stochastic approximation \citep{robbins1951stochastic} guarantees that if step sizes $\eta_t$ satisfy $\sum_t \eta_t = \infty$ and $\sum_t \eta_t^2 < \infty$, the discrete iterates converge almost surely to the stable equilibrium points of the ODE (Equation~\ref{eq:grad_flow} in our case). Assuming a sufficiently small constant $\eta$ and bounded variance, the system remains bounded within the basin and converges to a neighborhood of $\mH_{trap}$.

    \item We are interested in the correctness of the converged state. Since the trajectory converges to $\mH_{trap}$, the correctness of the final state is:
    \begin{align}
        \lim_{t \rightarrow \infty} J_{corr}(\mH) = J_{corr}(\lim_{t \rightarrow \infty} \mH) = J_{corr}(\mH_{trap}).
    \end{align}\\
    By the premise of the theorem, we have $J_{corr}(\mH_{trap}) < \epsilon$. Therefore, the optimization process drives the latent thought vectors to a fixed point $\mH_{trap}$ where the model is maximally confident ($\nabla_{\mH} J_{conf}(\mH_{trap}) = 0$) but the expected correctness is negligible ($J_{corr}(\mH_{trap}) < \epsilon$). The system is mathematically trapped in a high-confidence, low-accuracy state.
\end{enumerate}

\section{Generalization to Commonsense and Symbolic Reasoning}
\label{appendix:generalization}

In this section, we present the comprehensive experimental results supporting the generalization capabilities of LTPO discussed in Section~\ref{generalization}. To assess whether the benefits of LTPO extend beyond mathematical reasoning, we evaluated our framework on two distinct tasks: StrategyQA \citep{geva2021strategyqa} for commonsense reasoning and Date Understanding \citep{srivastava2023bigbench} from BIG-benchmark for symbolic reasoning.

The results, summarized in Table~\ref{tab:other_domain}, demonstrate that LTPO consistently outperforms both the Zero-Shot CoT baseline and the training-based SoftCoT baseline across diverse domains.

\paragraph{Commonsense reasoning.} StrategyQA requires models to infer implicit reasoning steps to answer boolean questions (e.g., ``Did aristotle use a laptop?''). LTPO demonstrates robust generalization to semantic reasoning tasks. On Qwen-2.5-7B-Instruct, LTPO achieves 73.09\% accuracy, surpassing the Zero-Shot CoT baseline of 72.63\%. Similarly, on LLaMA-3.1-8B-Instruct, LTPO improves upon the baseline (71.49\%) to reach 72.18\%. This indicates that the test-time optimization of latent thoughts maintains robust performance on natural language logic, providing consistent gains even on tasks distinct from mathematical reasoning.

\paragraph{Symbolic reasoning.} The Date Understanding task involves logical manipulation of time intervals (e.g., calculating relative dates). LTPO delivers its most significant non-mathematical improvements on this symbolic task. With Qwen-2.5-7B-Instruct, LTPO achieves 76.40\% accuracy, outperforming Zero-Shot CoT (64.40\%) by a substantial margin of +12.0\%. It also surpasses SoftCoT (68.40\%) by +8.0\%. This trend holds for LLaMA-3.1-8B-Instruct, where LTPO (62.00\%) outperforms both Zero-Shot CoT (55.60\%) and SoftCoT (58.40\%).

These experiments highlight a critical advantage of LTPO: adaptability. While training-based latent reasoning methods like SoftCoT can be brittle or underperform even with domain-specific training, LTPO leverage the model's intrinsic confidence landscape to discover optimal reasoning paths for each specific instance. This makes LTPO a highly robust framework capable of enhancing reasoning across mathematical, symbolic, and commonsense domains.

\begin{table}[h]
\caption{Performance of LTPO vs. SoftCoT vs. Zero-Shot CoT on StrategyQA (commonsense) and DU (symbolic) across 2 models, reported in accuracy (\%). ``DU'' indicates the Date Understanding dataset. The optimal results are in bold and the suboptimal ones are underlined.}
\label{tab:other_domain}
\centering
\begin{NiceTabular}{llcc|c}
\toprule
\multirow{2}{*}{\textbf{Model}} & \multirow{2}{*}{\textbf{Method}} & \textbf{Commonsense} & \textbf{Symbolic} & \multirow{2}{*}{\textbf{Avg.}} \\\cline{3-4} & & \textbf{StrategyQA} & \textbf{DU} & \\ \midrule
\midrule

\multirow{3}{*}{\begin{tabular}[c]{@{}l@{}}LLaMA-3.1-8B-\\Instruct\end{tabular}}
& Zero-Shot CoT & \underline{71.49} & 55.60 & \underline{63.55} \\
& SoftCoT & 67.34 & \underline{58.40} & 62.87 \\
& \textbf{LTPO (Ours)} & \textbf{72.18} & \textbf{62.00} & \textbf{67.09} \\ \midrule

\multirow{3}{*}{\begin{tabular}[c]{@{}l@{}}Qwen-2.5-7B-\\Instruct\end{tabular}}
& Zero-Shot CoT & \underline{72.63} & 64.40 & \underline{68.52} \\
& SoftCoT & 62.75 & \underline{68.40} & 65.58 \\
& \textbf{LTPO (Ours)} & \textbf{73.09} & \textbf{76.40} & \textbf{74.75} \\ \bottomrule

\end{NiceTabular}
\end{table}

\section{Analysis of Optimization Overhead vs. Generation Savings}
\label{appendix:inference_analysis}

In Section~\ref{inference_efficiency}, we reported that LTPO is faster than Zero-Shot CoT on challenging benchmarks, which may seem counter-intuitive given that LTPO introduces an iterative optimization loop at test time. Here, we provide a granular breakdown of the inference time to explain this phenomenon.

The total inference time consists of two parts: the optimization phase (unique to LTPO) and the final answer generation phase (common to all methods).
\begin{itemize}
    \item \textbf{Optimization Overhead:} The optimization loop in LTPO is computationally efficient because it does not involve autoregressive decoding. For the AIME2024 benchmark with $T=20$ steps, the optimization stage takes approximately \textbf{1.86 seconds} per problem.
    \item \textbf{Generation Savings:} The majority of the computational cost for reasoning tasks lies in the autoregressive generation of the final answer. We observe that optimizing latent thought tokens allows the model to reach the correct solution with a much more concise textual output.
\end{itemize}

Table~\ref{tab:token_len} compares the average generated token length between Zero-Shot CoT and LTPO on the AIME2024 benchmark. Zero-Shot CoT tends to generate long, winding reasoning chains (averaging 2,886 tokens). In contrast, LTPO, guided by the optimized latent thoughts, generates significantly shorter paths to the solution (averaging 1,420 tokens).

\begin{table}[h]
\caption{Comparison of average optimization time, generated token length, and total inference time per problem on the AIME2024 benchmark using LLaMA-3.1-8B-Instruct. The optimal results are in bold.}
\label{tab:token_len}
\centering
\begin{NiceTabular}{l|ccc}
\toprule
\textbf{Method} & \textbf{Opt. Time (s)} & \textbf{Avg. \# Tokens} & \textbf{Total Time (s)} \\ \midrule
\midrule

Zero-Shot CoT & \textbf{0.00} & 2886 & 62.59 \\
\textbf{LTPO (Ours)} & 1.86 & \textbf{1420} & \textbf{31.80} \\
\bottomrule

\end{NiceTabular}
\end{table}

This reduction in generation length (near 50\% reduction) far outweighs the minor cost of the optimization loop, resulting in a net speedup. This suggests that LTPO effectively compresses the necessary ``thinking'' into the latent vector space, reducing the need for verbose explicit reasoning.

\begin{table}[t]
\centering
\caption{Qualitative example of LTPO's confidence-correctness divergence. The model assigns a higher confidence-based reward to a fluent but incorrect reasoning path than to the correct solution, illustrating a key failure mode. The flawed intermediate step is marked with red color.}
\label{tab:case_study}
\resizebox{\linewidth}{!}{
\begin{tabular}{p{\linewidth}}
\toprule
\textcolor{Turquoise4}{\textbf{QUESTION:}}\\\\

What is the least positive integer multiple of 30 that can be written with only the digits 0 and 2? \\\\

\textcolor{Turquoise4}{\textbf{GROUND TRUTH:}}\\\\

2220
\\\\

\textcolor{Red3}{\textbf{CONFIDENT BUT INCORRECT ANSWER:}}
\\\\

\#\# Step 1:  To find the least positive integer multiple of 30 that can be written with only the digits 0 and 2, we need to consider the properties of multiples of 30. A multiple of 30 must be divisible by both 2 and 3.

\#\# Step 2:  Since the number must be written with only the digits 0 and 2, it must be an even number, making it divisible by 2. Therefore, we only need to focus on making it divisible by 3.

\#\# Step 3:  To be divisible by 3, the sum of the digits of the number must be divisible by 3. The smallest possible number that meets this condition and is written with only the digits 0 and 2 is 222, as 2 + 2 + 2 = 6, which is divisible by 3.

\#\# Step 4:  However, 222 is not a multiple of 30. To make it a multiple of 30, we need to multiply it by the smallest factor that will make it divisible by 10 (to make it a multiple of 10) and then by 3 (to maintain divisibility by 3).

\#\# Step 5:  \textcolor{Red3}{Multiplying 222 by 2 will make it divisible by 10, resulting in 444}. However, 444 is not a multiple of 3. We need to find the next multiple of 3 that is also a multiple of 10.

\#\# Step 6:  The next multiple of 3 that is also a multiple of 10 is 666, as 6 + 6 + 6 = 18, which is divisible by 3. However, we can find a smaller multiple by considering the properties of multiples of 30.

\#\# Step 7:  A multiple of 30 can be expressed as 30n, where n is a positive integer. To find the smallest multiple of 30 that can be written with only the digits 0 and 2, we need to find the smallest n such that 30n is a multiple of 10 and has only the digits 0 and 2.

\#\# Step 8:  The smallest n that satisfies these conditions is n = 4, as 30 * 4 = 120, which is a multiple of 10 and has only the digits 0 and 2.

\#\# Step 9:  Therefore, the least positive integer multiple of 30 that can be written with only the digits 0 and 2 is 120.

\textbackslash boxed\{120\}
\\\\

\textcolor{Green4}{\textbf{CORRECT ANSWER WITH LOWER CONFIDENCE}:}
\\\\

\#\# Step 1:  To find the least positive integer multiple of 30 that can be written with only the digits 0 and 2, we need to consider the properties of multiples of 30. A multiple of 30 must be divisible by both 2 and 3.

\#\# Step 2:  Since the number must be written with only the digits 0 and 2, it must be an even number, making it divisible by 2. Therefore, we only need to focus on making it divisible by 3.

\#\# Step 3:  To be divisible by 3, the sum of the digits of the number must be divisible by 3. The sum of the digits 0 and 2 is 2, which is not divisible by 3.

\#\# Step 4:  To make the sum of the digits divisible by 3, we need to add more digits that will make the total sum divisible by 3. The smallest number that can be formed using only the digits 0 and 2 and is divisible by 3 is 222.

\#\# Step 5:  However, 222 is not a multiple of 30. To make it a multiple of 30, we need to multiply it by 3, resulting in 666. But 666 is not a multiple of 30 either.

\#\# Step 6:  We need to find the smallest multiple of 222 that is also a multiple of 10 (since 30 = 2 * 3 * 5). The smallest multiple of 222 that is also a multiple of 10 is 2220.

\#\# Step 7:  Therefore, the least positive integer multiple of 30 that can be written with only the digits 0 and 2 is 2220.

\textbackslash boxed\{2220\}
\\
\bottomrule
\end{tabular}}
\end{table}

\section{Detailed Analysis of Latent Space Stability}
\label{appendix:latent_tokens_analysis}

In Section~\ref{impact_of_num_thought_tokens} and Figure~\ref{fig:linecharts} (left), we observed that LTPO maintains robust performance as the number of thought tokens ($K$) increases from 1 to 16, whereas baseline methods like SoftCoT degrade significantly. This stability raises a fundamental question: does the addition of thought tokens provide additional learning signals, or does the method merely ignore them? Here, we analyze the mechanics behind this robustness, distinguishing between static projection and dynamic optimization.

\subsection{Dynamic Optimization vs. Static Projection}

The divergence in performance stability between LTPO and SoftCoT stems from how latent thoughts are generated and utilized:

\paragraph{Static Projection (SoftCoT).} This approach relies on a static linear projection module, which is trained offline, to map the assistant model's thoughts into a fixed number of latent vectors. As the number of thought tokens increases, the static projection module struggles to produce high-quality representations without introducing noise. This results in the brittleness observed in Figure~\ref{fig:linecharts}, where SoftCoT degrades sharply beyond $K = 4$.

\paragraph{Dynamic Optimization (LTPO).} In contrast, LTPO treats the $K$ latent thought tokens as dynamic parameters to be optimized. We do not rely on a fixed mapping. Instead, the online policy gradient updates (Equation~\ref{eq:grad_ascent}) actively refine all $K$ vectors for each specific problem instance. This allows the model to ``prune'' non-useful directions in the latent space and align the vectors to maximize the intrinsic confidence reward, regardless of the initialization or count.

\subsection{The Latent ``Workspace'' Hypothesis}

We posit that LTPO treats the sequence of latent thought tokens as a dynamic ``workspace''. The stability of accuracy across varying $K$ demonstrates the method's ability to effectively manage this workspace capacity:

\paragraph{Adaptability.} For a given problem, LTPO can find the best way to use the allotted tokens. If a simple problem only requires minimal latent reasoning (e.g., equivalent to 1 or 2 thought tokens), LTPO will effectively optimize 2 tokens and the remaining 14 tokens will be refined to a state that does not interfere with the prediction.

\paragraph{Robustness to Redundancy.} Unlike static methods where redundant tokens introduce noise that disrupts the reasoning chain, LTPO's dynamic optimization ensures that additional capacity is either utilized for complex reasoning or neutralized.

Therefore, the stability of LTPO is not an indication that additional tokens lack signal. Rather, it validates that our test-time optimization effectively utilizes the allotted latent capacity, whether small or large, to find a high-confidence reasoning path without suffering from the degradation typical of static projection methods.

\section{Prompt Templates}
\label{appendix:prompt_templates}

In this section, we release the examples of AIME2024 for reference. Zero-Shot CoT-Unk and LTPO share the same prompt template, which is closely based on the one used by SoftCoT. We use reserved special tokens as the initial latent thought tokens.

In the following examples (Table~\ref{tab:prompt_example_1} and Table~\ref{tab:prompt_example_2}), content under the ``Answer'' section is the reasoning generated by the LLM, and the content under the ``Model Prediction'' section is the final numerical answer that is extracted by our parser.

\section{The Use of Large Language Models}

We acknowledge the use of the LLM, Gemini-2.5-Pro \citep{google2025gemini}, to help us polish the writing of this paper.

\begin{table}[t]
\centering
\caption{Prompt Example 1 for LTPO on AIME2024.}
\label{tab:prompt_example_1}
\resizebox{\linewidth}{!}{
\begin{tabular}{p{\linewidth}}
\toprule
\textbf{Input:}\\

Solve the following math problem efficiently and clearly:

- For simple problems (2 steps or fewer):

Provide a concise solution with minimal description.

- For complex problems (3 steps or more):

Use this step-by-step format:

\#\# Step 1: [Brief calculations]

\#\# Step 2: [Brief calculations]

...

IMPORTANT: Regardless of the approach, you MUST always put your final answer within \$\textbackslash boxed\{\{\}\}\$.

PROBLEM: Jen enters a lottery by picking $4$ distinct numbers from $S=\{1,2,3,\cdots,9,10\}.$ $4$ numbers are randomly chosen from $S.$ She wins a prize if at least two of her numbers were $2$ of the randomly chosen numbers, and wins the grand prize if all four of her numbers were the randomly chosen numbers. The probability of her winning the grand prize given that she won a prize is $\tfrac{m}{n}$ where $m$ and $n$ are relatively prime positive integers. Find $m+n$.

There are 8 special tokens that contain compressed latent reasoning information 
that might be useful for your reasoning.

If these tokens are useful for your case, you can use them as reference. If these tokens are not useful 
for your case, you can ignore them and focus back to solving the problem.

Here are the 8 special tokens: \texttt{[UNK]}\texttt{[UNK]}\texttt{[UNK]}\texttt{[UNK]}\texttt{[UNK]}\texttt{[UNK]}\texttt{[UNK]}\texttt{[UNK]}
\\\\

\textbf{Answer:}\\

\#\# Step 1: Calculate the total number of ways to choose 4 numbers from S.
The total number of ways to choose 4 numbers from S is given by the combination formula: $\binom{10}{4} = \frac{10!}{4!(10-4)!} = 210$.

\#\# Step 2: Calculate the number of ways to choose 4 numbers from S such that at least 2 of them match the randomly chosen numbers.
We can break this down into two cases: exactly 2 matches and exactly 3 or 4 matches.

\#\# Step 3: Calculate the number of ways to choose exactly 2 matches.
The number of ways to choose 2 numbers from the 4 randomly chosen numbers is $\binom{4}{2} = 6$. The number of ways to choose 2 numbers from the remaining 6 numbers in S is $\binom{6}{2} = 15$. Therefore, the total number of ways to choose exactly 2 matches is $6 \cdot 15 = 90$.

\#\# Step 4: Calculate the number of ways to choose exactly 3 or 4 matches.
The number of ways to choose 3 numbers from the 4 randomly chosen numbers is $\binom{4}{3} = 4$. The number of ways to choose 1 number from the remaining 6 numbers in S is $\binom{6}{1} = 6$. Therefore, the total number of ways to choose exactly 3 matches is $4 \cdot 6 = 24$. The number of ways to choose 4 numbers from the 4 randomly chosen numbers is $\binom{4}{4} = 1$. Therefore, the total number of ways to choose exactly 4 matches is $1$.

\#\# Step 5: Calculate the total number of ways to choose 4 numbers from S such that at least 2 of them match the randomly chosen numbers.
The total number of ways to choose 4 numbers from S such that at least 2 of them match the randomly chosen numbers is $90 + 24 + 1 = 115$.

\#\# Step 6: Calculate the probability of winning the grand prize given that she won a prize.
The probability of winning the grand prize given that she won a prize is $\frac{1}{115}$.

\#\# Step 7: Simplify the fraction to find m and n.
The fraction $\frac{1}{115}$ is already in its simplest form, so $m = 1$ and $n = 115$.

\#\# Step 8: Calculate the sum of m and n.
The sum of $m$ and $n$ is $1 + 115 = 116$.

The final answer is: \$\textbackslash boxed\{116\}\$
\\\\

\textbf{Model Prediction:}

116
\\\\

\textbf{Evaluate:}

\textcolor{Green4}{\textbf{Correct!}}

\\
\bottomrule
\end{tabular}}
\end{table}

\begin{table}[t]
\centering
\caption{Prompt Example 2 for LTPO on AIME2024.}
\label{tab:prompt_example_2}
\resizebox{\linewidth}{!}{
\begin{tabular}{p{\linewidth}}
\toprule
\textbf{Input:}\\

Solve the following math problem efficiently and clearly:

- For simple problems (2 steps or fewer):

Provide a concise solution with minimal description.

- For complex problems (3 steps or more):

Use this step-by-step format:

\#\# Step 1: [Brief calculations]

\#\# Step 2: [Brief calculations]

...

IMPORTANT: Regardless of the approach, you MUST always put your final answer within \$\textbackslash boxed\{\{\}\}\$.

PROBLEM: There exist real numbers $x$ and $y$, both greater than 1, such that $\log_x\left(y^x\right)=\log_y\left(x^{4y}\right)=10$. Find $xy$.

There are 8 special tokens that contain compressed latent reasoning information 
that might be useful for your reasoning.

If these tokens are useful for your case, you can use them as reference. If these tokens are not useful 
for your case, you can ignore them and focus back to solving the problem.

Here are the 8 special tokens: \texttt{[UNK]}\texttt{[UNK]}\texttt{[UNK]}\texttt{[UNK]}\texttt{[UNK]}\texttt{[UNK]}\texttt{[UNK]}\texttt{[UNK]}
\\\\

\textbf{Answer:}\\

\#\# Step 1:  We are given the equations $\log_x\left(y^x\right)=10$ and $\log_y\left(x^{4y}\right)=10$. We can rewrite these equations using the properties of logarithms: $\log_x\left(y^x\right) = x\log_x(y) = 10$ and $\log_y\left(x^{4y}\right) = 4y\log_y(x) = 10$.

\#\# Step 2:  Let's denote $\log_x(y)$ as $a$ and $\log_y(x)$ as $b$. Then we have the equations $xa = 10$ and $4yb = 10$.

\#\# Step 3:  From the first equation, we can express $a$ as $a = \frac{10}{x}$. Substituting this into the second equation, we get $4y\log_y(x) = 4yb = 10$, which simplifies to $b = \frac{10}{4y}$.

\#\# Step 4:  We also know that $b = \frac{1}{a}$. Substituting $a = \frac{10}{x}$, we get $b = \frac{1}{\frac{10}{x}} = \frac{x}{10}$.

\#\# Step 5:  Equating the two expressions for $b$, we have $\frac{x}{10} = \frac{10}{4y}$. Solving for $x$ in terms of $y$, we get $x = \frac{100}{4y} = \frac{25}{y}$.

\#\# Step 6:  Substituting $x = \frac{25}{y}$ into the equation $xa = 10$, we get $\frac{25}{y} \cdot a = 10$. Since $a = \frac{10}{x}$, we have $\frac{25}{y} \cdot \frac{10}{x} = 10$.

\#\# Step 7:  Simplifying the equation, we get $\frac{250}{xy} = 10$. Solving for $xy$, we have $xy = \frac{250}{10} = 25$.

The final answer is: \$\textbackslash boxed\{25\}\$
\\\\

\textbf{Model Prediction:}

25
\\\\

\textbf{Evaluate:}

\textcolor{Green4}{\textbf{Correct!}}

\\
\bottomrule
\end{tabular}}
\end{table}

\end{document}

%% file: math_commands.tex
\usepackage{amsmath,amsfonts,bm}

\def\eqref#1{equation~\ref{#1}}
\def\1{\bm{1}}

\def\vzero{{\bm{0}}}

\def\veps{{\bm{\epsilon}}}

\def\vsigma{{\bm{\sigma}}}

\def\vtheta{{\bm{\theta}}}
\def\va{{\bm{a}}}

\def\ve{{\bm{e}}}

\def\vh{{\bm{h}}}

\def\vx{{\bm{x}}}
\def\vy{{\bm{y}}}

\def\mA{{\bm{A}}}

\def\mH{{\bm{H}}}
\def\mI{{\bm{I}}}

\def\mV{{\bm{V}}}

\DeclareMathAlphabet{\mathsfit}{\encodingdefault}{\sfdefault}{m}{sl}
\SetMathAlphabet{\mathsfit}{bold}{\encodingdefault}{\sfdefault}{bx}{n}

\def\gB{{\mathcal{B}}}

\def\gH{{\mathcal{H}}}

\def\gM{{\mathcal{M}}}
\def\gN{{\mathcal{N}}}

\def\sR{{\mathbb{R}}}

\newcommand{\E}{\mathbb{E}}

%% file: iclr2026_conference.bib
@inproceedings{wei2022cot,
  author = {Wei, Jason and Wang, Xuezhi and Schuurmans, Dale and Bosma, Maarten and ichter, brian and Xia, Fei and Chi, Ed and Le, Quoc V and Zhou, Denny},
  booktitle = {Advances in Neural Information Processing Systems},
  editor = {S. Koyejo and S. Mohamed and A. Agarwal and D. Belgrave and K. Cho and A. Oh},
  pages = {24824--24837},
  publisher = {Curran Associates, Inc.},
  title = {Chain-of-Thought Prompting Elicits Reasoning in Large Language Models},
  url = {https://proceedings.neurips.cc/paper_files/paper/2022/file/9d5609613524ecf4f15af0f7b31abca4-Paper-Conference.pdf},
  volume = {35},
  year = {2022}
}

@article{hao2024training,
  title={Training Large Language Models to Reason in a Continuous Latent Space},
  author={Hao, Shibo and Sukhbaatar, Sainbayar and Su, DiJia and Li, Xian and Hu, Zhiting and Weston, Jason and Tian, Yuandong},
  journal={arXiv preprint arXiv:2412.06769},
  year={2024}
}

@inproceedings{Kojima2022zeroshot,
  author = {Kojima, Takeshi and Gu, Shixiang (Shane) and Reid, Machel and Matsuo, Yutaka and Iwasawa, Yusuke},
  booktitle = {Advances in Neural Information Processing Systems},
  editor = {S. Koyejo and S. Mohamed and A. Agarwal and D. Belgrave and K. Cho and A. Oh},
  pages = {22199--22213},
  publisher = {Curran Associates, Inc.},
  title = {Large Language Models are Zero-Shot Reasoners},
  url = {https://proceedings.neurips.cc/paper_files/paper/2022/file/8bb0d291acd4acf06ef112099c16f326-Paper-Conference.pdf},
  volume = {35},
  year = {2022}
}

@inproceedings{zhou2023leasttomost,
  title={Least-to-Most Prompting Enables Complex Reasoning in Large Language Models},
  author={Denny Zhou and Nathanael Sch{\"a}rli and Le Hou and Jason Wei and Nathan Scales and Xuezhi Wang and Dale Schuurmans and Claire Cui and Olivier Bousquet and Quoc V Le and Ed H. Chi},
  booktitle={The Eleventh International Conference on Learning Representations },
  year={2023},
  url={https://openreview.net/forum?id=WZH7099tgfM}
}

@misc{deng2024icot,
  title={From Explicit CoT to Implicit CoT: Learning to Internalize CoT Step by Step}, 
  author={Yuntian Deng and Yejin Choi and Stuart Shieber},
  year={2024},
  eprint={2405.14838},
  archivePrefix={arXiv},
  primaryClass={cs.CL},
  url={https://arxiv.org/abs/2405.14838}, 
}

@inproceedings{xu2025softcot,
  title = "{S}oft{C}o{T}: Soft Chain-of-Thought for Efficient Reasoning with {LLM}s",
  author = "Xu, Yige  and
    Guo, Xu  and
    Zeng, Zhiwei  and
    Miao, Chunyan",
  editor = "Che, Wanxiang  and
    Nabende, Joyce  and
    Shutova, Ekaterina  and
    Pilehvar, Mohammad Taher",
  booktitle = "Proceedings of the 63rd Annual Meeting of the Association for Computational Linguistics (Volume 1: Long Papers)",
  month = jul,
  year = "2025",
  address = "Vienna, Austria",
  publisher = "Association for Computational Linguistics",
  url = "https://aclanthology.org/2025.acl-long.1137/",
  doi = "10.18653/v1/2025.acl-long.1137",
  pages = "23336--23351",
  ISBN = "979-8-89176-251-0"
}

@inproceedings{ouyang2022rlhf,
  author = {Ouyang, Long and Wu, Jeffrey and Jiang, Xu and Almeida, Diogo and Wainwright, Carroll and Mishkin, Pamela and Zhang, Chong and Agarwal, Sandhini and Slama, Katarina and Ray, Alex and Schulman, John and Hilton, Jacob and Kelton, Fraser and Miller, Luke and Simens, Maddie and Askell, Amanda and Welinder, Peter and Christiano, Paul F and Leike, Jan and Lowe, Ryan},
  booktitle = {Advances in Neural Information Processing Systems},
  editor = {S. Koyejo and S. Mohamed and A. Agarwal and D. Belgrave and K. Cho and A. Oh},
  pages = {27730--27744},
  publisher = {Curran Associates, Inc.},
  title = {Training language models to follow instructions with human feedback},
  url = {https://proceedings.neurips.cc/paper_files/paper/2022/file/b1efde53be364a73914f58805a001731-Paper-Conference.pdf},
  volume = {35},
  year = {2022}
}

@misc{bai2022constitutionalaiharmlessnessai,
  title={Constitutional AI: Harmlessness from AI Feedback}, 
  author={Yuntao Bai and Saurav Kadavath and Sandipan Kundu and Amanda Askell and Jackson Kernion and Andy Jones and Anna Chen and Anna Goldie and Azalia Mirhoseini and Cameron McKinnon and Carol Chen and Catherine Olsson and Christopher Olah and Danny Hernandez and Dawn Drain and Deep Ganguli and Dustin Li and Eli Tran-Johnson and Ethan Perez and Jamie Kerr and Jared Mueller and Jeffrey Ladish and Joshua Landau and Kamal Ndousse and Kamile Lukosuite and Liane Lovitt and Michael Sellitto and Nelson Elhage and Nicholas Schiefer and Noemi Mercado and Nova DasSarma and Robert Lasenby and Robin Larson and Sam Ringer and Scott Johnston and Shauna Kravec and Sheer El Showk and Stanislav Fort and Tamera Lanham and Timothy Telleen-Lawton and Tom Conerly and Tom Henighan and Tristan Hume and Samuel R. Bowman and Zac Hatfield-Dodds and Ben Mann and Dario Amodei and Nicholas Joseph and Sam McCandlish and Tom Brown and Jared Kaplan},
  year={2022},
  eprint={2212.08073},
  archivePrefix={arXiv},
  primaryClass={cs.CL},
  url={https://arxiv.org/abs/2212.08073},
}

@misc{schulman2017ppo,
  title={Proximal Policy Optimization Algorithms}, 
  author={John Schulman and Filip Wolski and Prafulla Dhariwal and Alec Radford and Oleg Klimov},
  year={2017},
  eprint={1707.06347},
  archivePrefix={arXiv},
  primaryClass={cs.LG},
  url={https://arxiv.org/abs/1707.06347}, 
}

@inproceedings{Rafailov2023dpo,
  author = {Rafailov, Rafael and Sharma, Archit and Mitchell, Eric and Manning, Christopher D and Ermon, Stefano and Finn, Chelsea},
  booktitle = {Advances in Neural Information Processing Systems},
  editor = {A. Oh and T. Naumann and A. Globerson and K. Saenko and M. Hardt and S. Levine},
  pages = {53728--53741},
  publisher = {Curran Associates, Inc.},
  title = {Direct Preference Optimization: Your Language Model is Secretly a Reward Model},
  url = {https://proceedings.neurips.cc/paper_files/paper/2023/file/a85b405ed65c6477a4fe8302b5e06ce7-Paper-Conference.pdf},
  volume = {36},
  year = {2023}
}

@misc{shao2024grpo,
  title={DeepSeekMath: Pushing the Limits of Mathematical Reasoning in Open Language Models}, 
  author={Zhihong Shao and Peiyi Wang and Qihao Zhu and Runxin Xu and Junxiao Song and Xiao Bi and Haowei Zhang and Mingchuan Zhang and Y. K. Li and Y. Wu and Daya Guo},
  year={2024},
  eprint={2402.03300},
  archivePrefix={arXiv},
  primaryClass={cs.CL},
  url={https://arxiv.org/abs/2402.03300},
}

@misc{deepseekai2025deepseekr1,
  title={DeepSeek-R1: Incentivizing Reasoning Capability in LLMs via Reinforcement Learning}, 
  author={DeepSeek-AI and Daya Guo and Dejian Yang and Haowei Zhang and Junxiao Song and Ruoyu Zhang and Runxin Xu and Qihao Zhu and Shirong Ma and Peiyi Wang and Xiao Bi and Xiaokang Zhang and Xingkai Yu and Yu Wu and Z. F. Wu and Zhibin Gou and Zhihong Shao and Zhuoshu Li and Ziyi Gao and Aixin Liu and Bing Xue and Bingxuan Wang and Bochao Wu and Bei Feng and Chengda Lu and Chenggang Zhao and Chengqi Deng and Chenyu Zhang and Chong Ruan and Damai Dai and Deli Chen and Dongjie Ji and Erhang Li and Fangyun Lin and Fucong Dai and Fuli Luo and Guangbo Hao and Guanting Chen and Guowei Li and H. Zhang and Han Bao and Hanwei Xu and Haocheng Wang and Honghui Ding and Huajian Xin and Huazuo Gao and Hui Qu and Hui Li and Jianzhong Guo and Jiashi Li and Jiawei Wang and Jingchang Chen and Jingyang Yuan and Junjie Qiu and Junlong Li and J. L. Cai and Jiaqi Ni and Jian Liang and Jin Chen and Kai Dong and Kai Hu and Kaige Gao and Kang Guan and Kexin Huang and Kuai Yu and Lean Wang and Lecong Zhang and Liang Zhao and Litong Wang and Liyue Zhang and Lei Xu and Leyi Xia and Mingchuan Zhang and Minghua Zhang and Minghui Tang and Meng Li and Miaojun Wang and Mingming Li and Ning Tian and Panpan Huang and Peng Zhang and Qiancheng Wang and Qinyu Chen and Qiushi Du and Ruiqi Ge and Ruisong Zhang and Ruizhe Pan and Runji Wang and R. J. Chen and R. L. Jin and Ruyi Chen and Shanghao Lu and Shangyan Zhou and Shanhuang Chen and Shengfeng Ye and Shiyu Wang and Shuiping Yu and Shunfeng Zhou and Shuting Pan and S. S. Li and Shuang Zhou and Shaoqing Wu and Shengfeng Ye and Tao Yun and Tian Pei and Tianyu Sun and T. Wang and Wangding Zeng and Wanjia Zhao and Wen Liu and Wenfeng Liang and Wenjun Gao and Wenqin Yu and Wentao Zhang and W. L. Xiao and Wei An and Xiaodong Liu and Xiaohan Wang and Xiaokang Chen and Xiaotao Nie and Xin Cheng and Xin Liu and Xin Xie and Xingchao Liu and Xinyu Yang and Xinyuan Li and Xuecheng Su and Xuheng Lin and X. Q. Li and Xiangyue Jin and Xiaojin Shen and Xiaosha Chen and Xiaowen Sun and Xiaoxiang Wang and Xinnan Song and Xinyi Zhou and Xianzu Wang and Xinxia Shan and Y. K. Li and Y. Q. Wang and Y. X. Wei and Yang Zhang and Yanhong Xu and Yao Li and Yao Zhao and Yaofeng Sun and Yaohui Wang and Yi Yu and Yichao Zhang and Yifan Shi and Yiliang Xiong and Ying He and Yishi Piao and Yisong Wang and Yixuan Tan and Yiyang Ma and Yiyuan Liu and Yongqiang Guo and Yuan Ou and Yuduan Wang and Yue Gong and Yuheng Zou and Yujia He and Yunfan Xiong and Yuxiang Luo and Yuxiang You and Yuxuan Liu and Yuyang Zhou and Y. X. Zhu and Yanhong Xu and Yanping Huang and Yaohui Li and Yi Zheng and Yuchen Zhu and Yunxian Ma and Ying Tang and Yukun Zha and Yuting Yan and Z. Z. Ren and Zehui Ren and Zhangli Sha and Zhe Fu and Zhean Xu and Zhenda Xie and Zhengyan Zhang and Zhewen Hao and Zhicheng Ma and Zhigang Yan and Zhiyu Wu and Zihui Gu and Zijia Zhu and Zijun Liu and Zilin Li and Ziwei Xie and Ziyang Song and Zizheng Pan and Zhen Huang and Zhipeng Xu and Zhongyu Zhang and Zhen Zhang},
  year={2025},
  eprint={2501.12948},
  archivePrefix={arXiv},
  primaryClass={cs.CL},
  url={https://arxiv.org/abs/2501.12948},
}

@misc{yu2025dapo,
    title={DAPO: An Open-Source LLM Reinforcement Learning System at Scale}, 
    author={Qiying Yu and Zheng Zhang and Ruofei Zhu and Yufeng Yuan and Xiaochen Zuo and Yu Yue and Weinan Dai and Tiantian Fan and Gaohong Liu and Lingjun Liu and Xin Liu and Haibin Lin and Zhiqi Lin and Bole Ma and Guangming Sheng and Yuxuan Tong and Chi Zhang and Mofan Zhang and Wang Zhang and Hang Zhu and Jinhua Zhu and Jiaze Chen and Jiangjie Chen and Chengyi Wang and Hongli Yu and Yuxuan Song and Xiangpeng Wei and Hao Zhou and Jingjing Liu and Wei-Ying Ma and Ya-Qin Zhang and Lin Yan and Mu Qiao and Yonghui Wu and Mingxuan Wang},
    year={2025},
    eprint={2503.14476},
    archivePrefix={arXiv},
    primaryClass={cs.LG},
    url={https://arxiv.org/abs/2503.14476}, 
}

@misc{li2025latentseek,
  title={Seek in the Dark: Reasoning via Test-Time Instance-Level Policy Gradient in Latent Space}, 
  author={Hengli Li and Chenxi Li and Tong Wu and Xuekai Zhu and Yuxuan Wang and Zhaoxin Yu and Eric Hanchen Jiang and Song-Chun Zhu and Zixia Jia and Ying Nian Wu and Zilong Zheng},
  year={2025},
  eprint={2505.13308},
  archivePrefix={arXiv},
  primaryClass={cs.LG},
  url={https://arxiv.org/abs/2505.13308},
}

@misc{kang2025selfcertainty,
  title={Scalable Best-of-N Selection for Large Language Models via Self-Certainty}, 
  author={Zhewei Kang and Xuandong Zhao and Dawn Song},
  year={2025},
  eprint={2502.18581},
  archivePrefix={arXiv},
  primaryClass={cs.CL},
  url={https://arxiv.org/abs/2502.18581},
}

@misc{fu2025deepthinkconf,
  title={Deep Think with Confidence}, 
  author={Yichao Fu and Xuewei Wang and Yuandong Tian and Jiawei Zhao},
  year={2025},
  eprint={2508.15260},
  archivePrefix={arXiv},
  primaryClass={cs.LG},
  url={https://arxiv.org/abs/2508.15260},
}

@article{williams1992simple,
  title={Simple statistical gradient-following algorithms for connectionist reinforcement learning},
  author={Williams, Ronald J},
  journal={Machine learning},
  volume={8},
  pages={229--256},
  year={1992},
  publisher={Springer}
}

@misc{grattafiori2024llama3herdmodels,
  title={The Llama 3 Herd of Models}, 
  author={Aaron Grattafiori and Abhimanyu Dubey and Abhinav Jauhri and Abhinav Pandey and Abhishek Kadian and Ahmad Al-Dahle and Aiesha Letman and Akhil Mathur and Alan Schelten and Alex Vaughan and Amy Yang and Angela Fan and Anirudh Goyal and Anthony Hartshorn and Aobo Yang and Archi Mitra and Archie Sravankumar and Artem Korenev and Arthur Hinsvark and Arun Rao and Aston Zhang and Aurelien Rodriguez and Austen Gregerson and Ava Spataru and Baptiste Roziere and Bethany Biron and Binh Tang and Bobbie Chern and Charlotte Caucheteux and Chaya Nayak and Chloe Bi and Chris Marra and Chris McConnell and Christian Keller and Christophe Touret and Chunyang Wu and Corinne Wong and Cristian Canton Ferrer and Cyrus Nikolaidis and Damien Allonsius and Daniel Song and Danielle Pintz and Danny Livshits and Danny Wyatt and David Esiobu and Dhruv Choudhary and Dhruv Mahajan and Diego Garcia-Olano and Diego Perino and Dieuwke Hupkes and Egor Lakomkin and Ehab AlBadawy and Elina Lobanova and Emily Dinan and Eric Michael Smith and Filip Radenovic and Francisco Guzmán and Frank Zhang and Gabriel Synnaeve and Gabrielle Lee and Georgia Lewis Anderson and Govind Thattai and Graeme Nail and Gregoire Mialon and Guan Pang and Guillem Cucurell and Hailey Nguyen and Hannah Korevaar and Hu Xu and Hugo Touvron and Iliyan Zarov and Imanol Arrieta Ibarra and Isabel Kloumann and Ishan Misra and Ivan Evtimov and Jack Zhang and Jade Copet and Jaewon Lee and Jan Geffert and Jana Vranes and Jason Park and Jay Mahadeokar and Jeet Shah and Jelmer van der Linde and Jennifer Billock and Jenny Hong and Jenya Lee and Jeremy Fu and Jianfeng Chi and Jianyu Huang and Jiawen Liu and Jie Wang and Jiecao Yu and Joanna Bitton and Joe Spisak and Jongsoo Park and Joseph Rocca and Joshua Johnstun and Joshua Saxe and Junteng Jia and Kalyan Vasuden Alwala and Karthik Prasad and Kartikeya Upasani and Kate Plawiak and Ke Li and Kenneth Heafield and Kevin Stone and Khalid El-Arini and Krithika Iyer and Kshitiz Malik and Kuenley Chiu and Kunal Bhalla and Kushal Lakhotia and Lauren Rantala-Yeary and Laurens van der Maaten and Lawrence Chen and Liang Tan and Liz Jenkins and Louis Martin and Lovish Madaan and Lubo Malo and Lukas Blecher and Lukas Landzaat and Luke de Oliveira and Madeline Muzzi and Mahesh Pasupuleti and Mannat Singh and Manohar Paluri and Marcin Kardas and Maria Tsimpoukelli and Mathew Oldham and Mathieu Rita and Maya Pavlova and Melanie Kambadur and Mike Lewis and Min Si and Mitesh Kumar Singh and Mona Hassan and Naman Goyal and Narjes Torabi and Nikolay Bashlykov and Nikolay Bogoychev and Niladri Chatterji and Ning Zhang and Olivier Duchenne and Onur Çelebi and Patrick Alrassy and Pengchuan Zhang and Pengwei Li and Petar Vasic and Peter Weng and Prajjwal Bhargava and Pratik Dubal and Praveen Krishnan and Punit Singh Koura and Puxin Xu and Qing He and Qingxiao Dong and Ragavan Srinivasan and Raj Ganapathy and Ramon Calderer and Ricardo Silveira Cabral and Robert Stojnic and Roberta Raileanu and Rohan Maheswari and Rohit Girdhar and Rohit Patel and Romain Sauvestre and Ronnie Polidoro and Roshan Sumbaly and Ross Taylor and Ruan Silva and Rui Hou and Rui Wang and Saghar Hosseini and Sahana Chennabasappa and Sanjay Singh and Sean Bell and Seohyun Sonia Kim and Sergey Edunov and Shaoliang Nie and Sharan Narang and Sharath Raparthy and Sheng Shen and Shengye Wan and Shruti Bhosale and Shun Zhang and Simon Vandenhende and Soumya Batra and Spencer Whitman and Sten Sootla and Stephane Collot and Suchin Gururangan and Sydney Borodinsky and Tamar Herman and Tara Fowler and Tarek Sheasha and Thomas Georgiou and Thomas Scialom and Tobias Speckbacher and Todor Mihaylov and Tong Xiao and Ujjwal Karn and Vedanuj Goswami and Vibhor Gupta and Vignesh Ramanathan and Viktor Kerkez and Vincent Gonguet and Virginie Do and Vish Vogeti and Vítor Albiero and Vladan Petrovic and Weiwei Chu and Wenhan Xiong and Wenyin Fu and Whitney Meers and Xavier Martinet and Xiaodong Wang and Xiaofang Wang and Xiaoqing Ellen Tan and Xide Xia and Xinfeng Xie and Xuchao Jia and Xuewei Wang and Yaelle Goldschlag and Yashesh Gaur and Yasmine Babaei and Yi Wen and Yiwen Song and Yuchen Zhang and Yue Li and Yuning Mao and Zacharie Delpierre Coudert and Zheng Yan and Zhengxing Chen and Zoe Papakipos and Aaditya Singh and Aayushi Srivastava and Abha Jain and Adam Kelsey and Adam Shajnfeld and Adithya Gangidi and Adolfo Victoria and Ahuva Goldstand and Ajay Menon and Ajay Sharma and Alex Boesenberg and Alexei Baevski and Allie Feinstein and Amanda Kallet and Amit Sangani and Amos Teo and Anam Yunus and Andrei Lupu and Andres Alvarado and Andrew Caples and Andrew Gu and Andrew Ho and Andrew Poulton and Andrew Ryan and Ankit Ramchandani and Annie Dong and Annie Franco and Anuj Goyal and Aparajita Saraf and Arkabandhu Chowdhury and Ashley Gabriel and Ashwin Bharambe and Assaf Eisenman and Azadeh Yazdan and Beau James and Ben Maurer and Benjamin Leonhardi and Bernie Huang and Beth Loyd and Beto De Paola and Bhargavi Paranjape and Bing Liu and Bo Wu and Boyu Ni and Braden Hancock and Bram Wasti and Brandon Spence and Brani Stojkovic and Brian Gamido and Britt Montalvo and Carl Parker and Carly Burton and Catalina Mejia and Ce Liu and Changhan Wang and Changkyu Kim and Chao Zhou and Chester Hu and Ching-Hsiang Chu and Chris Cai and Chris Tindal and Christoph Feichtenhofer and Cynthia Gao and Damon Civin and Dana Beaty and Daniel Kreymer and Daniel Li and David Adkins and David Xu and Davide Testuggine and Delia David and Devi Parikh and Diana Liskovich and Didem Foss and Dingkang Wang and Duc Le and Dustin Holland and Edward Dowling and Eissa Jamil and Elaine Montgomery and Eleonora Presani and Emily Hahn and Emily Wood and Eric-Tuan Le and Erik Brinkman and Esteban Arcaute and Evan Dunbar and Evan Smothers and Fei Sun and Felix Kreuk and Feng Tian and Filippos Kokkinos and Firat Ozgenel and Francesco Caggioni and Frank Kanayet and Frank Seide and Gabriela Medina Florez and Gabriella Schwarz and Gada Badeer and Georgia Swee and Gil Halpern and Grant Herman and Grigory Sizov and Guangyi and Zhang and Guna Lakshminarayanan and Hakan Inan and Hamid Shojanazeri and Han Zou and Hannah Wang and Hanwen Zha and Haroun Habeeb and Harrison Rudolph and Helen Suk and Henry Aspegren and Hunter Goldman and Hongyuan Zhan and Ibrahim Damlaj and Igor Molybog and Igor Tufanov and Ilias Leontiadis and Irina-Elena Veliche and Itai Gat and Jake Weissman and James Geboski and James Kohli and Janice Lam and Japhet Asher and Jean-Baptiste Gaya and Jeff Marcus and Jeff Tang and Jennifer Chan and Jenny Zhen and Jeremy Reizenstein and Jeremy Teboul and Jessica Zhong and Jian Jin and Jingyi Yang and Joe Cummings and Jon Carvill and Jon Shepard and Jonathan McPhie and Jonathan Torres and Josh Ginsburg and Junjie Wang and Kai Wu and Kam Hou U and Karan Saxena and Kartikay Khandelwal and Katayoun Zand and Kathy Matosich and Kaushik Veeraraghavan and Kelly Michelena and Keqian Li and Kiran Jagadeesh and Kun Huang and Kunal Chawla and Kyle Huang and Lailin Chen and Lakshya Garg and Lavender A and Leandro Silva and Lee Bell and Lei Zhang and Liangpeng Guo and Licheng Yu and Liron Moshkovich and Luca Wehrstedt and Madian Khabsa and Manav Avalani and Manish Bhatt and Martynas Mankus and Matan Hasson and Matthew Lennie and Matthias Reso and Maxim Groshev and Maxim Naumov and Maya Lathi and Meghan Keneally and Miao Liu and Michael L. Seltzer and Michal Valko and Michelle Restrepo and Mihir Patel and Mik Vyatskov and Mikayel Samvelyan and Mike Clark and Mike Macey and Mike Wang and Miquel Jubert Hermoso and Mo Metanat and Mohammad Rastegari and Munish Bansal and Nandhini Santhanam and Natascha Parks and Natasha White and Navyata Bawa and Nayan Singhal and Nick Egebo and Nicolas Usunier and Nikhil Mehta and Nikolay Pavlovich Laptev and Ning Dong and Norman Cheng and Oleg Chernoguz and Olivia Hart and Omkar Salpekar and Ozlem Kalinli and Parkin Kent and Parth Parekh and Paul Saab and Pavan Balaji and Pedro Rittner and Philip Bontrager and Pierre Roux and Piotr Dollar and Polina Zvyagina and Prashant Ratanchandani and Pritish Yuvraj and Qian Liang and Rachad Alao and Rachel Rodriguez and Rafi Ayub and Raghotham Murthy and Raghu Nayani and Rahul Mitra and Rangaprabhu Parthasarathy and Raymond Li and Rebekkah Hogan and Robin Battey and Rocky Wang and Russ Howes and Ruty Rinott and Sachin Mehta and Sachin Siby and Sai Jayesh Bondu and Samyak Datta and Sara Chugh and Sara Hunt and Sargun Dhillon and Sasha Sidorov and Satadru Pan and Saurabh Mahajan and Saurabh Verma and Seiji Yamamoto and Sharadh Ramaswamy and Shaun Lindsay and Shaun Lindsay and Sheng Feng and Shenghao Lin and Shengxin Cindy Zha and Shishir Patil and Shiva Shankar and Shuqiang Zhang and Shuqiang Zhang and Sinong Wang and Sneha Agarwal and Soji Sajuyigbe and Soumith Chintala and Stephanie Max and Stephen Chen and Steve Kehoe and Steve Satterfield and Sudarshan Govindaprasad and Sumit Gupta and Summer Deng and Sungmin Cho and Sunny Virk and Suraj Subramanian and Sy Choudhury and Sydney Goldman and Tal Remez and Tamar Glaser and Tamara Best and Thilo Koehler and Thomas Robinson and Tianhe Li and Tianjun Zhang and Tim Matthews and Timothy Chou and Tzook Shaked and Varun Vontimitta and Victoria Ajayi and Victoria Montanez and Vijai Mohan and Vinay Satish Kumar and Vishal Mangla and Vlad Ionescu and Vlad Poenaru and Vlad Tiberiu Mihailescu and Vladimir Ivanov and Wei Li and Wenchen Wang and Wenwen Jiang and Wes Bouaziz and Will Constable and Xiaocheng Tang and Xiaojian Wu and Xiaolan Wang and Xilun Wu and Xinbo Gao and Yaniv Kleinman and Yanjun Chen and Ye Hu and Ye Jia and Ye Qi and Yenda Li and Yilin Zhang and Ying Zhang and Yossi Adi and Youngjin Nam and Yu and Wang and Yu Zhao and Yuchen Hao and Yundi Qian and Yunlu Li and Yuzi He and Zach Rait and Zachary DeVito and Zef Rosnbrick and Zhaoduo Wen and Zhenyu Yang and Zhiwei Zhao and Zhiyu Ma},
  year={2024},
  eprint={2407.21783},
  archivePrefix={arXiv},
  primaryClass={cs.AI},
  url={https://arxiv.org/abs/2407.21783},
}

@misc{yang2024qwen2technicalreport,
  title={Qwen2 Technical Report}, 
  author={An Yang and Baosong Yang and Binyuan Hui and Bo Zheng and Bowen Yu and Chang Zhou and Chengpeng Li and Chengyuan Li and Dayiheng Liu and Fei Huang and Guanting Dong and Haoran Wei and Huan Lin and Jialong Tang and Jialin Wang and Jian Yang and Jianhong Tu and Jianwei Zhang and Jianxin Ma and Jianxin Yang and Jin Xu and Jingren Zhou and Jinze Bai and Jinzheng He and Junyang Lin and Kai Dang and Keming Lu and Keqin Chen and Kexin Yang and Mei Li and Mingfeng Xue and Na Ni and Pei Zhang and Peng Wang and Ru Peng and Rui Men and Ruize Gao and Runji Lin and Shijie Wang and Shuai Bai and Sinan Tan and Tianhang Zhu and Tianhao Li and Tianyu Liu and Wenbin Ge and Xiaodong Deng and Xiaohuan Zhou and Xingzhang Ren and Xinyu Zhang and Xipin Wei and Xuancheng Ren and Xuejing Liu and Yang Fan and Yang Yao and Yichang Zhang and Yu Wan and Yunfei Chu and Yuqiong Liu and Zeyu Cui and Zhenru Zhang and Zhifang Guo and Zhihao Fan},
  year={2024},
  eprint={2407.10671},
  archivePrefix={arXiv},
  primaryClass={cs.CL},
  url={https://arxiv.org/abs/2407.10671}, 
}

@misc{yang2025qwen3technicalreport,
  title={Qwen3 Technical Report}, 
  author={An Yang and Anfeng Li and Baosong Yang and Beichen Zhang and Binyuan Hui and Bo Zheng and Bowen Yu and Chang Gao and Chengen Huang and Chenxu Lv and Chujie Zheng and Dayiheng Liu and Fan Zhou and Fei Huang and Feng Hu and Hao Ge and Haoran Wei and Huan Lin and Jialong Tang and Jian Yang and Jianhong Tu and Jianwei Zhang and Jianxin Yang and Jiaxi Yang and Jing Zhou and Jingren Zhou and Junyang Lin and Kai Dang and Keqin Bao and Kexin Yang and Le Yu and Lianghao Deng and Mei Li and Mingfeng Xue and Mingze Li and Pei Zhang and Peng Wang and Qin Zhu and Rui Men and Ruize Gao and Shixuan Liu and Shuang Luo and Tianhao Li and Tianyi Tang and Wenbiao Yin and Xingzhang Ren and Xinyu Wang and Xinyu Zhang and Xuancheng Ren and Yang Fan and Yang Su and Yichang Zhang and Yinger Zhang and Yu Wan and Yuqiong Liu and Zekun Wang and Zeyu Cui and Zhenru Zhang and Zhipeng Zhou and Zihan Qiu},
  year={2025},
  eprint={2505.09388},
  archivePrefix={arXiv},
  primaryClass={cs.CL},
  url={https://arxiv.org/abs/2505.09388}, 
}

@misc{liu20251bllmsurpass405b,
  title={Can 1B LLM Surpass 405B LLM? Rethinking Compute-Optimal Test-Time Scaling}, 
  author={Runze Liu and Junqi Gao and Jian Zhao and Kaiyan Zhang and Xiu Li and Biqing Qi and Wanli Ouyang and Bowen Zhou},
  year={2025},
  eprint={2502.06703},
  archivePrefix={arXiv},
  primaryClass={cs.CL},
  url={https://arxiv.org/abs/2502.06703}, 
}

@misc{cobbe2021gsm8k,
  title={Training Verifiers to Solve Math Word Problems}, 
  author={Karl Cobbe and Vineet Kosaraju and Mohammad Bavarian and Mark Chen and Heewoo Jun and Lukasz Kaiser and Matthias Plappert and Jerry Tworek and Jacob Hilton and Reiichiro Nakano and Christopher Hesse and John Schulman},
  year={2021},
  eprint={2110.14168},
  archivePrefix={arXiv},
  primaryClass={cs.LG},
  url={https://arxiv.org/abs/2110.14168}, 
}

@inproceedings{hendrycks2021math500,
  title={Measuring Mathematical Problem Solving With the {MATH} Dataset},
  author={Dan Hendrycks and Collin Burns and Saurav Kadavath and Akul Arora and Steven Basart and Eric Tang and Dawn Song and Jacob Steinhardt},
  booktitle={Thirty-fifth Conference on Neural Information Processing Systems Datasets and Benchmarks Track (Round 2)},
  year={2021},
  url={https://openreview.net/forum?id=7Bywt2mQsCe}
}

@misc{aime2024,
  title = {AIME 2024},
  url = {https://huggingface.co/datasets/Maxwell-Jia/AIME_2024},
  author = {{Mathematical Association of America}},
  year = {2024}
}

@misc{aime2025,
  title = {AIME 2025},
  url = {https://huggingface.co/datasets/opencompass/AIME2025},
  author = {{Mathematical Association of America}},
  year = {2025}
}

@inproceedings{miao2020asdiv,
  title = "A Diverse Corpus for Evaluating and Developing {E}nglish Math Word Problem Solvers",
  author = "Miao, Shen-yun  and  Liang, Chao-Chun  and Su, Keh-Yih",
  editor = "Jurafsky, Dan  and  Chai, Joyce  and  Schluter, Natalie  and  Tetreault, Joel",
  booktitle = "Proceedings of the 58th Annual Meeting of the Association for Computational Linguistics",
  month = jul,
  year = "2020",
  address = "Online",
  publisher = "Association for Computational Linguistics",
  url = "https://aclanthology.org/2020.acl-main.92/",
  doi = "10.18653/v1/2020.acl-main.92",
  pages = "975--984"
}

@misc{google2025gemini,
  title={Gemini 2.5: Our most intelligent AI model},
  author={Google},
  year={2025},
  url={https://blog.google/technology/google-deepmind/gemini-model-thinking-updates-march-2025/}
}

@misc{xai2025grok4,
  title={Grok 4},
  author={xAI},
  year={2025},
  url={https://x.ai/news/grok-4}
}

@misc{openai2025gpt5,
  title={Introducing GPT-5},
  author={OpenAI},
  year={2025},
  url={https://openai.com/index/introducing-gpt-5/}
}

@misc{balunović2025matharena,
  title={MathArena: Evaluating LLMs on Uncontaminated Math Competitions}, 
  author={Mislav Balunović and Jasper Dekoninck and Ivo Petrov and Nikola Jovanović and Martin Vechev},
  year={2025},
  eprint={2505.23281},
  archivePrefix={arXiv},
  primaryClass={cs.AI},
  url={https://arxiv.org/abs/2505.23281}, 
}

@inproceedings{hu2022lora,
  title={Lo{RA}: Low-Rank Adaptation of Large Language Models},
  author={Edward J Hu and yelong shen and Phillip Wallis and Zeyuan Allen-Zhu and Yuanzhi Li and Shean Wang and Lu Wang and Weizhu Chen},
  booktitle={International Conference on Learning Representations},
  year={2022},
  url={https://openreview.net/forum?id=nZeVKeeFYf9}
}

@inproceedings{xu2025genius,
    title = "Genius: A Generalizable and Purely Unsupervised Self-Training Framework For Advanced Reasoning",
    author = "Xu, Fangzhi  and
      Yan, Hang  and
      Ma, Chang  and
      Zhao, Haiteng  and
      Sun, Qiushi  and
      Cheng, Kanzhi  and
      He, Junxian  and
      Liu, Jun  and
      Wu, Zhiyong",
    editor = "Che, Wanxiang  and
      Nabende, Joyce  and
      Shutova, Ekaterina  and
      Pilehvar, Mohammad Taher",
    booktitle = "Proceedings of the 63rd Annual Meeting of the Association for Computational Linguistics (Volume 1: Long Papers)",
    month = jul,
    year = "2025",
    address = "Vienna, Austria",
    publisher = "Association for Computational Linguistics",
    url = "https://aclanthology.org/2025.acl-long.644/",
    doi = "10.18653/v1/2025.acl-long.644",
    pages = "13153--13167",
    ISBN = "979-8-89176-251-0"
}

@inproceedings{zeng2025simplerlzoo,
    title={Simple{RL}-Zoo: Investigating and Taming Zero Reinforcement Learning for Open Base Models in the Wild},
    author={Weihao Zeng and Yuzhen Huang and Qian Liu and Wei Liu and Keqing He and Zejun MA and Junxian He},
    booktitle={Second Conference on Language Modeling},
    year={2025},
    url={https://openreview.net/forum?id=vSMCBUgrQj}
}

@article{geva2021strategyqa,
    title = "Did Aristotle Use a Laptop? A Question Answering Benchmark with Implicit Reasoning Strategies",
    author = "Geva, Mor  and
      Khashabi, Daniel  and
      Segal, Elad  and
      Khot, Tushar  and
      Roth, Dan  and
      Berant, Jonathan",
    editor = "Roark, Brian  and
      Nenkova, Ani",
    journal = "Transactions of the Association for Computational Linguistics",
    volume = "9",
    year = "2021",
    address = "Cambridge, MA",
    publisher = "MIT Press",
    url = "https://aclanthology.org/2021.tacl-1.21/",
    doi = "10.1162/tacl_a_00370",
    pages = "346--361"
}

@article{srivastava2023bigbench,
title={Beyond the Imitation Game: Quantifying and extrapolating the capabilities of language models},
author={Aarohi Srivastava and Abhinav Rastogi and Abhishek Rao and Abu Awal Md Shoeb and Abubakar Abid and Adam Fisch and Adam R. Brown and Adam Santoro and Aditya Gupta and Adri{\`a} Garriga-Alonso and Agnieszka Kluska and Aitor Lewkowycz and Akshat Agarwal and Alethea Power and Alex Ray and Alex Warstadt and Alexander W. Kocurek and Ali Safaya and Ali Tazarv and Alice Xiang and Alicia Parrish and Allen Nie and Aman Hussain and Amanda Askell and Amanda Dsouza and Ambrose Slone and Ameet Rahane and Anantharaman S. Iyer and Anders Johan Andreassen and Andrea Madotto and Andrea Santilli and Andreas Stuhlm{\"u}ller and Andrew M. Dai and Andrew La and Andrew Kyle Lampinen and Andy Zou and Angela Jiang and Angelica Chen and Anh Vuong and Animesh Gupta and Anna Gottardi and Antonio Norelli and Anu Venkatesh and Arash Gholamidavoodi and Arfa Tabassum and Arul Menezes and Arun Kirubarajan and Asher Mullokandov and Ashish Sabharwal and Austin Herrick and Avia Efrat and Aykut Erdem and Ayla Karaka{\c{s}} and B. Ryan Roberts and Bao Sheng Loe and Barret Zoph and Bart{\l}omiej Bojanowski and Batuhan {\"O}zyurt and Behnam Hedayatnia and Behnam Neyshabur and Benjamin Inden and Benno Stein and Berk Ekmekci and Bill Yuchen Lin and Blake Howald and Bryan Orinion and Cameron Diao and Cameron Dour and Catherine Stinson and Cedrick Argueta and Cesar Ferri and Chandan Singh and Charles Rathkopf and Chenlin Meng and Chitta Baral and Chiyu Wu and Chris Callison-Burch and Christopher Waites and Christian Voigt and Christopher D Manning and Christopher Potts and Cindy Ramirez and Clara E. Rivera and Clemencia Siro and Colin Raffel and Courtney Ashcraft and Cristina Garbacea and Damien Sileo and Dan Garrette and Dan Hendrycks and Dan Kilman and Dan Roth and C. Daniel Freeman and Daniel Khashabi and Daniel Levy and Daniel Mosegu{\'\i} Gonz{\'a}lez and Danielle Perszyk and Danny Hernandez and Danqi Chen and Daphne Ippolito and Dar Gilboa and David Dohan and David Drakard and David Jurgens and Debajyoti Datta and Deep Ganguli and Denis Emelin and Denis Kleyko and Deniz Yuret and Derek Chen and Derek Tam and Dieuwke Hupkes and Diganta Misra and Dilyar Buzan and Dimitri Coelho Mollo and Diyi Yang and Dong-Ho Lee and Dylan Schrader and Ekaterina Shutova and Ekin Dogus Cubuk and Elad Segal and Eleanor Hagerman and Elizabeth Barnes and Elizabeth Donoway and Ellie Pavlick and Emanuele Rodol{\`a} and Emma Lam and Eric Chu and Eric Tang and Erkut Erdem and Ernie Chang and Ethan A Chi and Ethan Dyer and Ethan Jerzak and Ethan Kim and Eunice Engefu Manyasi and Evgenii Zheltonozhskii and Fanyue Xia and Fatemeh Siar and Fernando Mart{\'\i}nez-Plumed and Francesca Happ{\'e} and Francois Chollet and Frieda Rong and Gaurav Mishra and Genta Indra Winata and Gerard de Melo and Germ{\`a}n Kruszewski and Giambattista Parascandolo and Giorgio Mariani and Gloria Xinyue Wang and Gonzalo Jaimovitch-Lopez and Gregor Betz and Guy Gur-Ari and Hana Galijasevic and Hannah Kim and Hannah Rashkin and Hannaneh Hajishirzi and Harsh Mehta and Hayden Bogar and Henry Francis Anthony Shevlin and Hinrich Schuetze and Hiromu Yakura and Hongming Zhang and Hugh Mee Wong and Ian Ng and Isaac Noble and Jaap Jumelet and Jack Geissinger and Jackson Kernion and Jacob Hilton and Jaehoon Lee and Jaime Fern{\'a}ndez Fisac and James B Simon and James Koppel and James Zheng and James Zou and Jan Kocon and Jana Thompson and Janelle Wingfield and Jared Kaplan and Jarema Radom and Jascha Sohl-Dickstein and Jason Phang and Jason Wei and Jason Yosinski and Jekaterina Novikova and Jelle Bosscher and Jennifer Marsh and Jeremy Kim and Jeroen Taal and Jesse Engel and Jesujoba Alabi and Jiacheng Xu and Jiaming Song and Jillian Tang and Joan Waweru and John Burden and John Miller and John U. Balis and Jonathan Batchelder and Jonathan Berant and J{\"o}rg Frohberg and Jos Rozen and Jose Hernandez-Orallo and Joseph Boudeman and Joseph Guerr and Joseph Jones and Joshua B. Tenenbaum and Joshua S. Rule and Joyce Chua and Kamil Kanclerz and Karen Livescu and Karl Krauth and Karthik Gopalakrishnan and Katerina Ignatyeva and Katja Markert and Kaustubh Dhole and Kevin Gimpel and Kevin Omondi and Kory Wallace Mathewson and Kristen Chiafullo and Ksenia Shkaruta and Kumar Shridhar and Kyle McDonell and Kyle Richardson and Laria Reynolds and Leo Gao and Li Zhang and Liam Dugan and Lianhui Qin and Lidia Contreras-Ochando and Louis-Philippe Morency and Luca Moschella and Lucas Lam and Lucy Noble and Ludwig Schmidt and Luheng He and Luis Oliveros-Col{\'o}n and Luke Metz and L{\"u}tfi Kerem Senel and Maarten Bosma and Maarten Sap and Maartje Ter Hoeve and Maheen Farooqi and Manaal Faruqui and Mantas Mazeika and Marco Baturan and Marco Marelli and Marco Maru and Maria Jose Ramirez-Quintana and Marie Tolkiehn and Mario Giulianelli and Martha Lewis and Martin Potthast and Matthew L Leavitt and Matthias Hagen and M{\'a}ty{\'a}s Schubert and Medina Orduna Baitemirova and Melody Arnaud and Melvin McElrath and Michael Andrew Yee and Michael Cohen and Michael Gu and Michael Ivanitskiy and Michael Starritt and Michael Strube and Micha{\l} Sw{\k{e}}drowski and Michele Bevilacqua and Michihiro Yasunaga and Mihir Kale and Mike Cain and Mimee Xu and Mirac Suzgun and Mitch Walker and Mo Tiwari and Mohit Bansal and Moin Aminnaseri and Mor Geva and Mozhdeh Gheini and Mukund Varma T and Nanyun Peng and Nathan Andrew Chi and Nayeon Lee and Neta Gur-Ari Krakover and Nicholas Cameron and Nicholas Roberts and Nick Doiron and Nicole Martinez and Nikita Nangia and Niklas Deckers and Niklas Muennighoff and Nitish Shirish Keskar and Niveditha S. Iyer and Noah Constant and Noah Fiedel and Nuan Wen and Oliver Zhang and Omar Agha and Omar Elbaghdadi and Omer Levy and Owain Evans and Pablo Antonio Moreno Casares and Parth Doshi and Pascale Fung and Paul Pu Liang and Paul Vicol and Pegah Alipoormolabashi and Peiyuan Liao and Percy Liang and Peter W Chang and Peter Eckersley and Phu Mon Htut and Pinyu Hwang and Piotr Mi{\l}kowski and Piyush Patil and Pouya Pezeshkpour and Priti Oli and Qiaozhu Mei and Qing Lyu and Qinlang Chen and Rabin Banjade and Rachel Etta Rudolph and Raefer Gabriel and Rahel Habacker and Ramon Risco and Rapha{\"e}l Milli{\`e}re and Rhythm Garg and Richard Barnes and Rif A. Saurous and Riku Arakawa and Robbe Raymaekers and Robert Frank and Rohan Sikand and Roman Novak and Roman Sitelew and Ronan Le Bras and Rosanne Liu and Rowan Jacobs and Rui Zhang and Russ Salakhutdinov and Ryan Andrew Chi and Seungjae Ryan Lee and Ryan Stovall and Ryan Teehan and Rylan Yang and Sahib Singh and Saif M. Mohammad and Sajant Anand and Sam Dillavou and Sam Shleifer and Sam Wiseman and Samuel Gruetter and Samuel R. Bowman and Samuel Stern Schoenholz and Sanghyun Han and Sanjeev Kwatra and Sarah A. Rous and Sarik Ghazarian and Sayan Ghosh and Sean Casey and Sebastian Bischoff and Sebastian Gehrmann and Sebastian Schuster and Sepideh Sadeghi and Shadi Hamdan and Sharon Zhou and Shashank Srivastava and Sherry Shi and Shikhar Singh and Shima Asaadi and Shixiang Shane Gu and Shubh Pachchigar and Shubham Toshniwal and Shyam Upadhyay and Shyamolima Shammie Debnath and Siamak Shakeri and Simon Thormeyer and Simone Melzi and Siva Reddy and Sneha Priscilla Makini and Soo-Hwan Lee and Spencer Torene and Sriharsha Hatwar and Stanislas Dehaene and Stefan Divic and Stefano Ermon and Stella Biderman and Stephanie Lin and Stephen Prasad and Steven Piantadosi and Stuart Shieber and Summer Misherghi and Svetlana Kiritchenko and Swaroop Mishra and Tal Linzen and Tal Schuster and Tao Li and Tao Yu and Tariq Ali and Tatsunori Hashimoto and Te-Lin Wu and Th{\'e}o Desbordes and Theodore Rothschild and Thomas Phan and Tianle Wang and Tiberius Nkinyili and Timo Schick and Timofei Kornev and Titus Tunduny and Tobias Gerstenberg and Trenton Chang and Trishala Neeraj and Tushar Khot and Tyler Shultz and Uri Shaham and Vedant Misra and Vera Demberg and Victoria Nyamai and Vikas Raunak and Vinay Venkatesh Ramasesh and vinay uday prabhu and Vishakh Padmakumar and Vivek Srikumar and William Fedus and William Saunders and William Zhang and Wout Vossen and Xiang Ren and Xiaoyu Tong and Xinran Zhao and Xinyi Wu and Xudong Shen and Yadollah Yaghoobzadeh and Yair Lakretz and Yangqiu Song and Yasaman Bahri and Yejin Choi and Yichi Yang and Sophie Hao and Yifu Chen and Yonatan Belinkov and Yu Hou and Yufang Hou and Yuntao Bai and Zachary Seid and Zhuoye Zhao and Zijian Wang and Zijie J. Wang and Zirui Wang and Ziyi Wu},
journal={Transactions on Machine Learning Research},
issn={2835-8856},
year={2023},
url={https://openreview.net/forum?id=uyTL5Bvosj},
note={Featured Certification}
}

@article{lyapunov1992stability,
author = {A. M. Lyapunov},
title = {The general problem of the stability of motion},
journal = {International Journal of Control},
volume = {55},
number = {3},
pages = {531--534},
year = {1992},
publisher = {Taylor \& Francis},
doi = {10.1080/00207179208934253},
URL = { https://doi.org/10.1080/00207179208934253},
eprint = {https://doi.org/10.1080/00207179208934253}
}

@article{robbins1951stochastic,
author = {Herbert Robbins and Sutton Monro},
title = {{A Stochastic Approximation Method}},
volume = {22},
journal = {The Annals of Mathematical Statistics},
number = {3},
publisher = {Institute of Mathematical Statistics},
pages = {400 -- 407},
year = {1951},
doi = {10.1214/aoms/1177729586},
URL = {https://doi.org/10.1214/aoms/1177729586}
}
